\documentclass[11pt]{article} 
\usepackage{booktabs}

\usepackage[letterpaper, left=1in, right=1in, top=0.95in, bottom=0.95in]{geometry}

\usepackage[hidelinks=true]{hyperref}
\usepackage{natbib}
\usepackage{amsthm}
\usepackage{graphicx}
\usepackage{amsmath}
\usepackage{amssymb}
\usepackage{amsfonts}
\usepackage{caption}
\usepackage{cleveref}

\usepackage[utf8]{inputenc} 

\newcommand{\p}{\mathbb{P}}

\newcommand{\E}{\mathbb{E}}

\newcommand{\reals}{\mathbb{R}}

\newcommand{\cF}{\mathcal{F}}

\newcommand{\cH}{\mathcal{H}}
\newcommand{\cR}{\mathcal{R}}

\newcommand{\cX}{\mathcal{X}}

\DeclareMathOperator*{\argmin}{arg\,min}

\newcommand{\hY}{\hat{Y}}

\newcommand{\indep}{\perp \!\!\!\perp}

\newcommand{\hs}[1]{\hspace{#1}}

\newcommand{\Cov}{\mathrm{Cov}}
\newcommand{\Var}{\mathrm{Var}}

\usepackage{algorithm}
\usepackage{algpseudocode}
\usepackage{bbm}
\usepackage{subcaption}

\theoremstyle{plain}
\newtheorem{theorem}{Theorem}[section]

\newtheorem{lemma}[theorem]{Lemma}
\newtheorem{corollary}[theorem]{Corollary}
\theoremstyle{definition}
\newtheorem{definition}[theorem]{Definition}

\theoremstyle{remark}

\usepackage{setspace}

\title{Integrating Expert Judgment and Algorithmic Decision Making: \\ An Indistinguishability Framework}

\author{%
Rohan Alur$^{1,}$\thanks{Correspondence to ralur@mit.edu. Code is available at \href{https://github.com/ralur/integrating-expert-judgment}{https://github.com/ralur/integrating-expert-judgment.}}\qquad 
Loren Laine\(^{2}\)  \qquad Darrick K. Li\(^{2}\) \\
\vspace{5pt} \\
Dennis Shung\(^{2}\)  \qquad Manish Raghavan$^{1}$  \qquad  Devavrat Shah$^1$ 
\vspace{14pt} 
\\
\small{$^1$Massachusetts Institute of Technology, $^2$Yale School of Medicine}  
} 

\date{}

\begin{document}

\maketitle 

\begin{abstract}
We introduce a novel framework for human-AI collaboration in prediction and decision tasks. Our approach leverages human judgment to distinguish inputs which are \emph{algorithmically indistinguishable}, or ``look the same" to any feasible predictive algorithm. We argue that this framing clarifies the problem of human-AI collaboration in prediction and decision tasks, as experts often form judgments by drawing on information which is not encoded in an algorithm's training data. Algorithmic indistinguishability yields a natural test for assessing whether experts incorporate this kind of ``side information", and further provides a simple but principled method for selectively incorporating human feedback into algorithmic predictions. We show that this method provably improves the performance of any feasible algorithmic predictor and precisely quantify this improvement. We demonstrate the utility of our framework in a case study of emergency room triage decisions, where we find that although algorithmic risk scores are highly competitive with physicians, there is strong evidence that physician judgments provide signal which could not be replicated by \emph{any} predictive algorithm. This insight yields a range of natural decision rules which leverage the complementary strengths of human experts and predictive algorithms.

\end{abstract}


\maketitle

\section{Introduction}\label{sec: introduction}

Our work begins with the following puzzle: since at least 1954, it has been known that ``actuarial" predictions --- those made by algorithms, even very simple ones --- often outperform even expert human decision makers \citep{Meehl1954ClinicalVS}. Indeed, in 1989, the clinical psychologist Paul Meehl remarked that ``no controversy in social science...shows such a large body of qualitatively diverse studies coming out so uniformly in the same direction as this one" \citep{Meehl1986}. Nonetheless, and despite dramatic advances in machine learning over the intervening decades, human judgment continues to play a central role in most high-stakes prediction tasks.

One explanation for this apparent contradiction is that humans often have access to \emph{information} which is not encoded in the data available to predictive algorithms. For example, consider the problem of triage in the emergency room, where healthcare providers assess and prioritize patients for immediate care. While it is perhaps tempting to supplement clinicians' discretionary judgment with an algorithmic risk score -- and indeed, a large body of literature suggests that these algorithms may be more accurate than their human counterparts \citep{bias-productivity-2018, clinical-v-actuarial-1989, clinical-v-mechanical-2000, human-decisions-machine-predictions-2017, mechanical-vs-clinical-2013, graduate-admissions-1971, diagnosing-expertise-2017, diagnosing-physician-error-2019} -- predictive algorithms will typically only have access to structured data stored in a patient's electronic health records. In contrast, a physician has access to many additional modalities --- not least of which is the ability to directly examine the patient! Furthermore, physicians can draw on extensive background knowledge, formal training and intuitive judgments which are difficult to replicate in the classical statistical learning paradigm \citep{Kabrhel2005, Cook2009, Stolper2011}. Our work thus seeks to answer the following question:

\medskip
\begin{center}
\emph{When (and how) can human judgment improve the predictions of any learning algorithm?} \\
\end{center}

\medskip
Answering this question is of central importance to managers and policymakers. On one hand, investments in automation have the potential to dramatically improve both efficiency and accuracy relative to the status quo. On the other hand, in contexts where humans have access to information which is not encoded in an algorithm's training data, there is reason to believe that \emph{no} learning algorithm, no matter how sophisticated, can obviate the need for human expertise. Furthermore, there are many contexts in which decision makers cannot deploy arbitrarily sophisticated algorithms due to e.g., limited computational resources, training data, technical expertise or other logistical constraints.\footnote{For example, \cite{Marwaha2022} document the myriad practical challenges associated with deploying algorithmic risk scores and other digital tools in large hospital systems.} Thus, in deciding whether (and to what degree) to invest in automating a given task, the relevant question is often not whether humans outperform a particular algorithm \emph{on average}, but instead whether incorporating human judgment can improve the performance of the best algorithm which is feasible to deploy in a given setting.

\subsection{Case study: emergency room triage} \label{sec: case study}
To illustrate these concerns, consider the aforementioned problem of \emph{triage} for patients who present with acute gastrointestinal bleeding in the emergency room (we study this task in detail in \Cref{sec:experiments}). For each of these patients, the attending physician is tasked with deciding whether the bleeding is severe enough to warrant admission to the hospital. To aid physicians in making this determination efficiently, the \emph{Glasgow-Blatchford Score} (GBS) \citep{Blatchford2000-ce} is a validated algorithmic risk score that can be used as a highly sensitive indicator for whether a patient will require hospital-based interventions \citep{BARKUN2019, Laine2021-oe}. Nonetheless, we are interested in whether we can improve these algorithmic risk assessments by incorporating a second opinion from the physician, particularly because the physician's impression (or ``gestalt") may be informed by subjective factors which are not available to the GBS.

\textbf{Observationally indistinguishable patients.} A first heuristic, without making any assumptions about the construction of the GBS, is to ask whether a physician can distinguish between high and low risk patients whose electronic health records are \emph{identical}. Because the GBS is determined by a combination of existing data in the health record and additional pre-specified elements recorded by providers at presentation, this implies that all such patients will be assigned the same risk score. Indeed, this is true of any algorithmic risk score which uses only the electronic health record as input. Thus, a physician who can correctly distinguish between high and low risk patients within a group of these ``observationally indistinguishable" patients must necessarily be incorporating information which is not available to a predictive algorithm. 

We demonstrate the results of this exercise below on $2893$ patients who presented with acute gastrointestinal bleeding in the emergency room of a large academic hospital system (this represents a random $80\%$ subsample of our full dataset containing $3617$ patients; we hold out the remaining $20\%$ for validating experiments in \Cref{sec:experiments}). We group these patients into subsets which are \emph{observationally indistinguishable} on the basis of the nine discrete-valued inputs to the GBS. For example, the GBS uses a binary indicator for hepatic (liver) disease as one of its input features; thus, all patients within an observationally indistinguishable subset will share the same value of this feature. Within each of these subsets, we plot the Matthew's Correlation Coefficient\footnote{We use the Matthew's Correlation Coefficient (MCC) as a standard measure of binary classification accuracy \citep{advantages-mcc-2020}. The MCC is the usual Pearson Correlation Coefficient specialized to the case of binary-valued random variables.} between the physician's binary decision whether or not to hospitalize a given patient and a binary indicator for whether a patient suffered an adverse event (to streamline presentation, we defer additional discussion of this dataset, including the precise outcome definition and the potential selective labels issue that arises, to \Cref{sec:experiments}). These results are presented in \Cref{fig: cond mcc} below, where points above the x-axis indicate that the physician decisions are informative about a patient's true risk \emph{conditional} on the characteristics which serve as inputs to the GBS. 

\begin{figure}[htpb!]
\captionsetup{justification=centering}
\includegraphics[scale=.50]{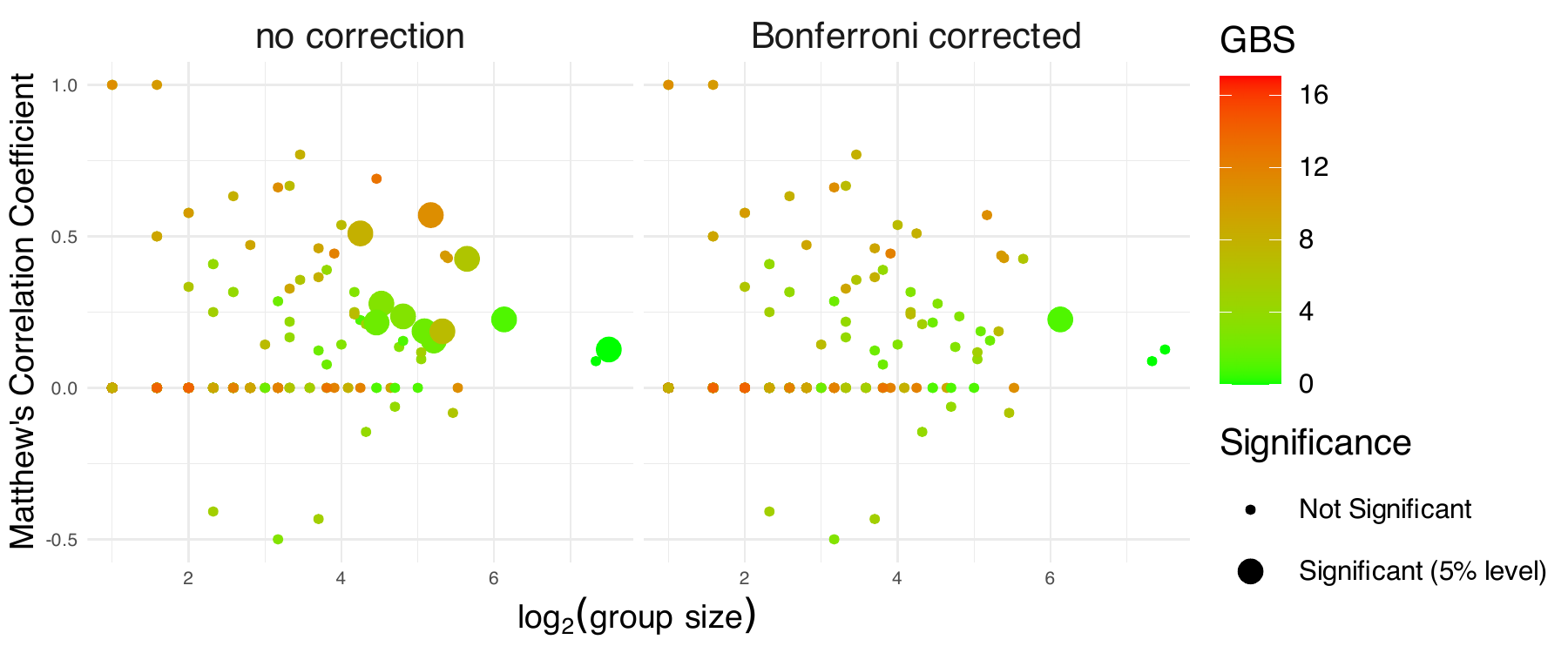}
    \centering
    \caption{The correlation between the physician's decision (hospitalize or discharge) and adverse outcomes within the $669$ observationally indistinguishable groups of patients. Large points are statistically significant at the $5\%$ level (estimated from $1000$ bootstrap replicates). The right panel applies a Bonferroni correction to account for multiple testing (i.e., large points are significantly different from $0$ at the $.05/669$ level). }

    \label{fig: cond mcc}
\end{figure}

\begin{figure}[htpb!]
\includegraphics[scale=.50]{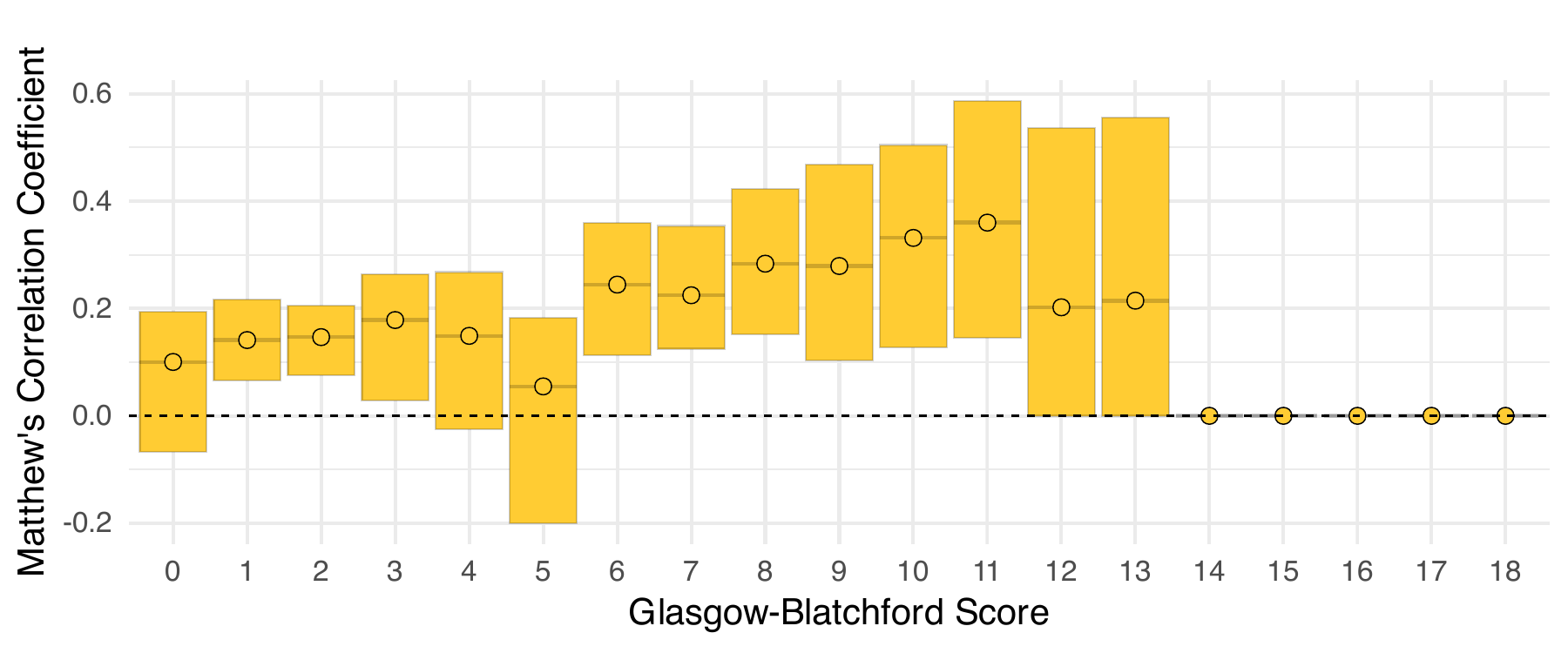}
\captionsetup{justification=centering}
    \centering
    \caption{The correlation between the physician's decision (hospitalize or discharge) and adverse outcomes within the level sets of the Glasgow-Blatchford Score. Point estimates are reported with $95\%$ Bonferroni Corrected confidence intervals (estimated from $1000$ bootstrap replicates).}
    \label{fig: by_gbs}
\end{figure}

This exercise provides suggestive evidence that physicians incorporate information which is not encoded in the nine inputs to the Glasgow-Blatchford Score. Indeed, the physician's judgment appears to provide additional value relative to \emph{any} predictor trained on these same inputs, as no function can distinguish patients who are observationally indistinguishable. Of course, \Cref{fig: cond mcc} highlights the key technical challenge: even in a setting with only nine discrete-valued features, most patients are `almost' unique; there are $669$ observationally indistinguishable subsets, and most of these contain just a handful of patients. Thus, while it is possible to show that physician judgments add predictive signal \emph{on average}, trying to ascertain \emph{which} patients would particularly benefit from a physician assessment leads to a severe multiple testing problem. This challenge is only compounded in settings where the data are continuous-valued and/or high-dimensional.
\newpage
\textbf{Algorithmically indistinguishable patients.} To begin to address these challenges, we now consider a much simpler problem: rather than asking whether physician judgment provides signal which cannot be communicated \emph{any} predictive algorithm, can we instead characterize the value of physician judgment relative to the Glasgow-Blatchford Score alone? To evaluate this, we simply stratify patients into groups which are assigned the same risk score by the GBS and again examine the conditional correlation between hospitalization decisions and adverse outcomes. We plot these below in \Cref{fig: by_gbs}.

As \Cref{fig: by_gbs} demonstrates, physicians' decision remain substantially predictive of adverse event risk even conditional on most values of the Glasgow-Blatchford Score. This is true both for high risk patients (GBS $\in [6, 11]$), and for patients with scores immediately above and below the recommended hospitalization threshold of $2$ \citep{BARKUN2019, Laine2021-oe}. Thus, we interpret this as demonstrating that physicians succeed in distinguishing patients who are indistinguishable on the basis of the GBS alone.

\Cref{fig: cond mcc} and \Cref{fig: by_gbs} can be interpreted as spanning two extremes of some spectrum, albeit one which we have yet to formally define. In \Cref{fig: cond mcc} we look to make strong, information-theoretic claims about the value of physician judgment with respect to \emph{any} algorithmic predictor, but lack statistical power to operationalize our approach. In \Cref{fig: by_gbs}, we instead test whether physicians incorporate information which is not communicated by the Glasgow-Blatchford Score alone, but cannot generalize our findings beyond this particular risk score. The remainder of our work seeks to interpolate between these two approaches to achieve the best of both worlds. We define a notion of \emph{algorithmic indistinguishability}, which smoothly characterizes which instances are ``effectively" identical from the perspective of some rich, user-defined class of feasible predictive algorithms, and show that this allows us to leverage the complementary strengths of humans and algorithms for both prediction and decision making. We elaborate on these contributions below.

\subsection{Contributions}

We propose a novel framework for human-AI collaboration in prediction and decision tasks. Our approach uses human judgment to refine predictions within sets of inputs which are \emph{algorithmically indistinguishable}, or ``look the same" to any feasible predictive algorithm. This class of feasible algorithms is domain-specific, and is the critical ingredient which parameterizes our approach. At one extreme, this class may contain only a single, fixed predictor; in this case, inputs which receive the same prediction (i.e., the level sets of the predictor) are algorithmically indistinguishable. We generalize this intuition to richer, infinitely large classes of predictive models, and argue that this framing helps clarify the role of human judgment in complementing algorithmic predictions.

In \Cref{sec: preliminaries}, we provide technical preliminaries and an overview of our methodology. In \Cref{sec: results} we provide our main technical results, which show that indistinguishability allows a decision maker to selectively incorporate human feedback only when it provably improves on the best feasible predictive model (and precisely quantify this improvement). We also show that, in the important special case of binary classification, indistinguishability can enable a stronger information-theoretic characterization of whether human judgment provides signal which cannot be replicated by \emph{any} feasible algorithmic predictor. 

In \Cref{sec:experiments} we apply these results to the emergency room triage task discussed in the previous section, where we find that physician judgment provides predictive signal  which is complementary not just to the Glasglow-Blatchford Score, but also with respect to \emph{any} shallow (depth $\le 3$) regression tree. This provides an intuitive basis for decision making, collapsing the space of more than $5000$ unique patient characteristics into just $7$ ``indistinguishable" subsets. In particular, indistinguishability greatly simplifies the space of possible decision rules --- for each patient, whether to solicit physician input or defer to an algorithmic risk score --- and yields a range of attractive policies.

Our technical results have connections to \emph{multicalibration} \citep{multicalibration-2017} and \emph{omniprediction} \citep{omnipredictors}, and our approach is conceptually inspired by the notion of \emph{outcome indistinguishability} \citep{outcome-indistinguishability}. Our work extends the main result of \cite{omnipredictors} in the special case of squared error, which may be of independent interest. We provide a detailed discussion of this connection in \Cref{sec: omnipredictors comparison}, and elaborate on the relationship to other related work below.

\section{Related work}
\label{sec:related}

\textbf{The relative strengths of humans and algorithms.} Our work is motivated by large body of literature which studies the relative strengths of human judgment and algorithmic decision making \citep{bias-productivity-2018, clinical-v-actuarial-1989, clinical-v-mechanical-2000, mechanical-v-clinical} or identifies behavioral biases in decision making \citep{heuristics-biases-1974, process-performance-paradox-1991, discrimination-bail-decisions-2020, prediction-mistakes-2022}. More recent work also studies whether predictive algorithms can \emph{improve} expert decision making \citep{human-decisions-machine-predictions-2017, diagnosing-physician-error-2019, improving-decisions-2021, human-expertise-ai-2023}.

\textbf{Recommendations, deferral and complementarity.} One popular approach for incorporating human judgment into algorithmic predictions is by \emph{deferring} some instances to a human decision maker \citep{learning-to-defer-2018, algo-triage-2019, consistent-deferral-2020, multiple-experts-2021, differentiable-triage-2021, closed-deferral-pipelines-2022}. Other work studies contexts where human decision makers are free to override algorithmic recommendations \citep{case-for-hil-2020, human-centered-eval-2020, algorithmic-social-eng-2020, overcoming-aversion-2018, human-expertise-ai-2023}, which may suggest alternative design criteria for these algorithms \citep{accuracy-teamwork-2020, human-aligned-calibration-2023}. More generally, systems which achieve human-AI \emph{complementarity} (as defined in \Cref{sec: introduction}) have been previously studied in \citet{prediction-vs-judgment-2018, accuracy-teamwork-2020, learning-to-complement-2020, hai-complementarity-2022, bayesian-complementarity-2022, classification-under-assistance-2020, beyond-learning-to-defer-2022}.

\citet{taxonomy-complementarity-2022} develop a comprehensive taxonomy of this area, which generally takes the predictor as given, or learns a predictor which is optimized to complement a particular model of human decision making. Our work takes a different perspective, and instead seeks to characterize the complementary role of human judgment in contexts where a decision maker may choose from among a rich (often infinitely large) class of feasible predictive models.

\textbf{Performative prediction.} A recent line of work studies \emph{performative prediction} \citep{performative-prediction-2020}, or settings in which predictions influence future outcomes. For example, predicting the risk of adverse health outcomes may directly inform treatment decisions, which in turn affects future health outcomes. This can complicate the design and evaluation of predictive algorithms, and there is a growing literature which seeks to address these challenges \citep{Brown2020PerformativePI, MendlerDnner2020StochasticOF, Jagadeesan2022RegretMW, Kim2022MakingDU, MendlerDnner2022AnticipatingPB, harmful-prophecies-2023, Hardt2023PerformativePP, Zhao2023PerformativeTF}. Performativity is also closely related to the \emph{selective labels problem}, in which some historical outcomes are unobserved as a consequence of past human decisions \citep{selective-labels-2017}. Though these issues arise in many canonical human-AI collaboration tasks, we focus on standard supervised learning problems in which predictions do not causally affect the outcome of interest. These include e.g., weather prediction, stock price forecasting and many medical diagnosis and triage tasks. In particular, although a physician's decision may inform subsequent treatment decisions, it does not affect the contemporaneous presence, absence or severity of a given condition. More generally, our work can be applied to any ``prediction policy problem'', where accurate predictions can be translated into policy gains without explicitly modeling causality \citep{prediction-policy-2015}.

\textbf{Multicalibration, omnipredictors and boosting.} Our results draw on recent developments in learning theory, particularly work on \emph{omnipredictors} \citep{omnipredictors} and its connections to \emph{multicalibration}. \citet{outcome-indistinguishability} show that multicalibration is tightly connected to a cryptographic notion of indistinguishability, which serves as conceptual inspiration for our work.\citet{multicalibration-boosting-regression} provide an elegant boosting algorithm for learning multicalibrated partitions that we make use of in our experiments, and \citet{omniprediction-via-multicalibration} provide results which reveal tight connections between a related notion of ``swap agnostic learning'', multi-group fairness, omniprediction and outcome indistinguishability.

\section{Methodology and preliminaries}
\label{sec: preliminaries}

In this section we develop a simple method for selectively incorporating human feedback to improve algorithmic predictions. Our approach generalizes the intuition outlined in \Cref{sec: case study}, which focused on the use of physician judgment to distinguish patients who are ``observationally indistinguishable," or share the same representation in the algorithm's input space. This strategy is intuitively appealing, but, as discussed in \Cref{sec: case study}, it quickly becomes statistically intractable as the dimension of the input space grows. One way to deal with this limitation is to consider inputs which are ``close" (but not necessarily identical) to be indistinguishable, as suggested by \citet{auditing-expertise}. In this section we define a richer notion of \emph{algorithmic indistinguishability}, which instead characterizes conditions under which input are ``effectively" identical for the purposes of making algorithmic predictions. We show that this definition admits a conceptually simple method for incorporating human feedback only when it improves the performance of any feasible predictive algorithm while remaining statistically tractable. 

\textbf{Notation.} Let $X \in \cX$ be a random variable denoting the inputs (or ``features'') which are available for making algorithmic predictions about an outcome $Y \in [0, 1]$. Let $\hY \in [0, 1]$ be an expert's prediction of $Y$, and let $x, y, \hat{y}$ denote realizations of the corresponding random variables. Our approach is parameterized by a class of predictors $\cF$, which is some set of functions mapping $\cX$ to $[0, 1]$. We interpret $\cF$ as the class of predictive models which are relevant (or feasible to implement) for a given prediction task; we discuss this choice below. Broadly, we are interested in whether the expert prediction $\hY$ provides a predictive signal which cannot be extracted from $X$ by any $f \in \cF$.

\textbf{Choice of model class $\cF$.} For now we place no restrictions on $\cF$, but it's helpful to consider a concrete model class (e.g., a specific neural network architecture) from which, given some training data, one could derive a \emph{particular} model (e.g., via empirical risk minimization over $\cF$). The choice of $\cF$ could be guided by practical considerations; for example, some domains might require interpretable models (e.g., linear functions) or impose computational constraints at training and/or inference time. We might also simply believe that a certain architecture or functional form is well suited to the task of interest. In any case, we are interested in whether human judgment can provide information which is not conveyed by any model in this class, but are agnostic as to \emph{how} this is accomplished: an expert may have information which is not encoded in $X$, or be deploying a decision rule which is not in $\cF$ --- or both!

Another perspective is to interpret $\cF$ as modeling more abstract limitations on the expert's cognitive process. In particular, to model experts who are subject to ``bounded rationality'' \citep{models-of-man-1957, bounded-rationality-history-2005}, $\cF$ might be the set of functions which can be efficiently computed (e.g., by a circuit of limited complexity). In this case, an expert who provides a prediction which cannot be modeled by any $f \in \cF$ must have access to \emph{information} which is not present in the training data. We take the choice of $\cF$ as given, but emphasize that these two approaches yield qualitatively different insight about human expertise.

\textbf{Indistinguishability with respect to $\cF$.} Our approach will be to use human input to distinguish observations which are \emph{indistinguishable} to any predictor $f \in \cF$. We formalize this notion of indistinguishability as follows:

\begin{definition}[$\alpha$-Indistinguishable subset]\label{def: indistinguishable subset} For some $\alpha \ge 0$, a set $S \subseteq \cX$ is $\alpha$-indistinguishable with respect to a function class $\cF$ and target $Y$ if, for all $f \in \cF$,
\begin{equation}\label{eq: alpha-indistinguishable}
    \left|\Cov(f(X), Y \mid X \in S)\right| \le \alpha 
\end{equation}
\end{definition}

To interpret this definition, observe that the subset $S$ can be viewed as generalizing the intuition given in \Cref{sec: introduction} for grouping observationally indistinguishable inputs. In particular, rather than requiring that all $x \in S$ are exactly equal, \Cref{def: indistinguishable subset} requires that all members of $S$ effectively ``look the same" for the purposes of making algorithmic predictions about $Y$, as every $f \in \cF$ is only weakly related to the outcome within $S$. We now adopt the definition of a multicalibrated partition \citep{omnipredictors} as follows:

\begin{definition}[$\alpha$-Multicalibrated partition]\label{def: multicalibrated partition} For $K \geq 1$, $S_1 \dots S_K \subseteq \cX$ is an $\alpha$-multicalibrated partition with respect to $\cF$ and $Y$ if (1) $S_1 \dots S_K$ partitions $\cX$ and (2) each $S_k$ is $\alpha$-indistinguishable with respect to $\cF$ and $Y$.\footnote{This is closely related to $\alpha$-approximate multicalibration \citep{omnipredictors}, which requires that \Cref{def: indistinguishable subset} merely holds in expectation over the partition. We work with a stronger pointwise definition for clarity, but our results can also be interpreted as holding for the `typical' element of an $\alpha$-approximately multicalibrated partition.}
\end{definition}

\begin{figure}[h!]
\captionsetup{justification=centering}
    \centering
    \begin{subfigure}[b]{0.45\textwidth}
        \centering
        \includegraphics[width=\textwidth]{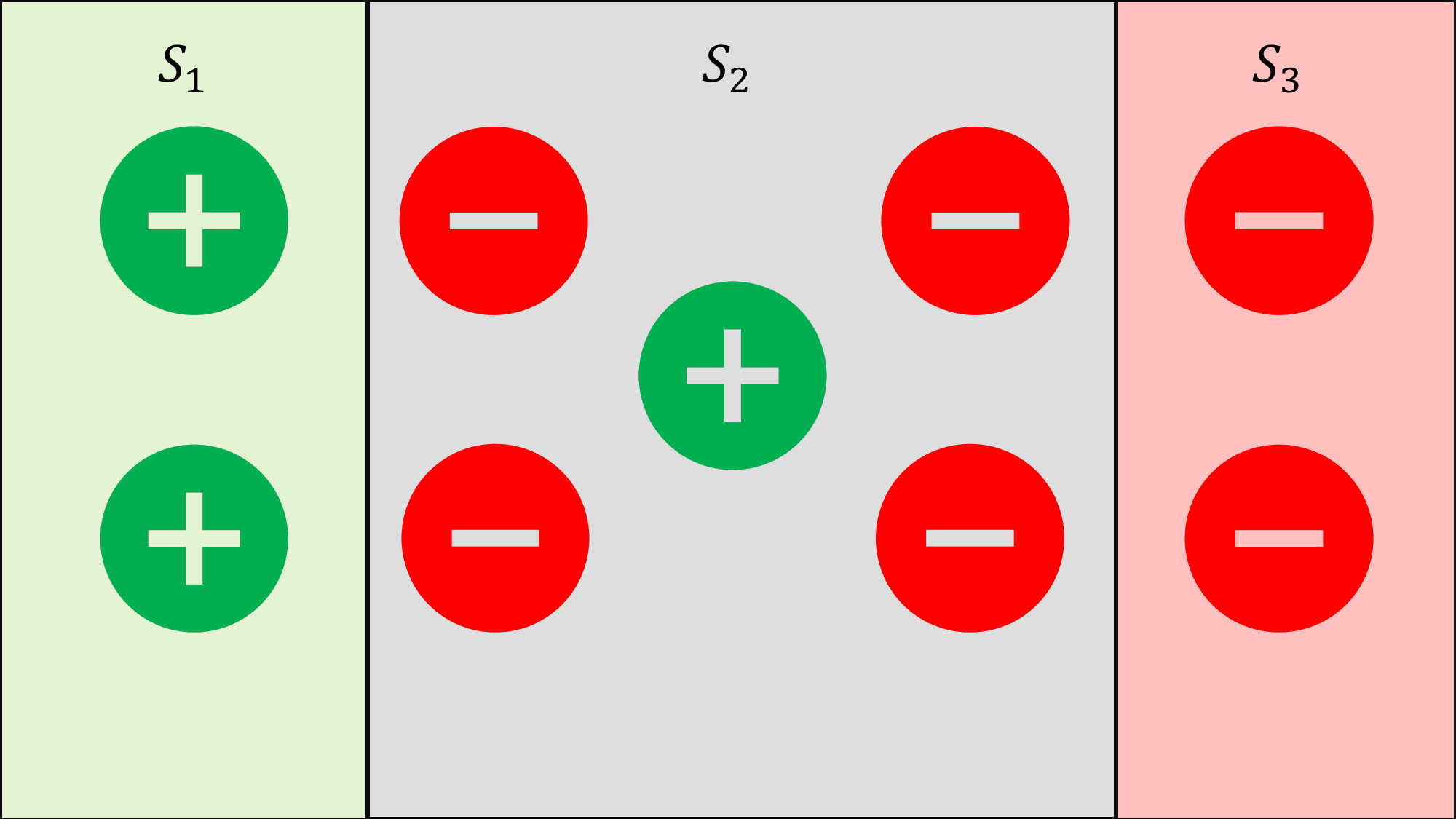}
        \caption{}
        \label{fig:subfig1}
    \end{subfigure}
    \hfill
    \begin{subfigure}[b]{0.45\textwidth}
        \centering
        \includegraphics[width=\textwidth]{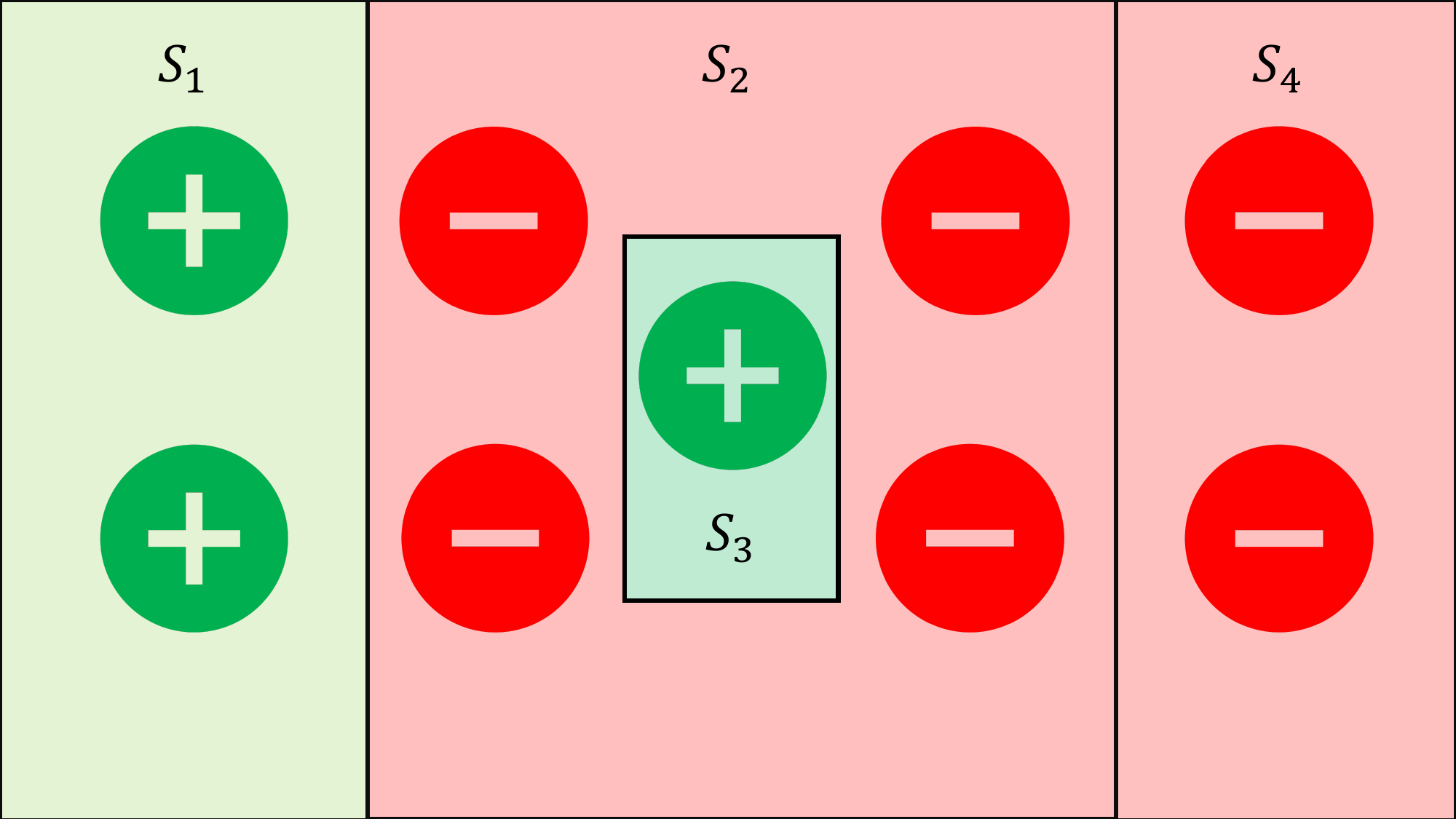}
        \caption{}
        \label{fig:subfig2}
    \end{subfigure}
    \caption{Partitions which are approximately multicalibrated with respect to the class of hyperplane classifiers. For illustration purposes, we consider the empirical distribution placing equal probability on each observation. In both panels, no hyperplane classifier has significant discriminatory power within each subset.}
    \label{fig:mc_stylized_example}
\end{figure}

Intuitively, the partition $\{S_k\}_{k \in [K]}$ ``extract[s] all the predictive power" from $\cF$ \citep{omnipredictors}. In particular, while knowing that an input $x$ lies in some subset $S_k$ may be highly informative for predicting $Y$ --- for example, it may be that $\E[Y \mid X = x] \approx \E[Y \mid X = x']$ for all $x, x' \in S_k$ --- no predictor $f \in \cF$ provides significant \emph{additional} signal within this subset. We provide a stylized example of such partitions in \Cref{fig:mc_stylized_example} above. 

It's not obvious that such partitions are feasible to compute, or even that they should exist. We'll show in \Cref{sec:learning partitions} however that a multicalibrated partition can be efficiently computed for many natural classes of functions. Where the relevant partition is clear from context, we use $\E_k[\cdot], \Var_k(\cdot), \Cov_k(\cdot, \cdot)$ to denote expectation, variance and covariance conditional on the event that $\{X \in S_k\}$. For a subset $S \subseteq \cX$, we use $\E_S[\cdot], \Var_S(\cdot)$ and $\Cov_S(\cdot, \cdot)$ analogously.

\textbf{Incorporating human judgment into predictions.} To incorporate human judgment into predictions, a natural heuristic is to first test whether the conditional covariance $\Cov_k(Y, \hY)$ is nonzero within some indistinguishable subset. Intuitively, this indicates that within $S_k$, the expert prediction is informative even though every algorithm $f \in \cF$ is not. This suggests the following approach: first, learn a partition which is multicalibrated with respect to $\cF$, and then use $\hY$ to predict $Y$ within each indistinguishable subset. We describe this procedure in \Cref{alg:meta_algo} below, where we define a univariate learning algorithm ${\cal A}$ as a procedure which takes one or more $(\hat{y}_i, y_i) \in [0,1]^2$ training observations and outputs a function which predicts $Y$ using $\hat{Y}$.  For example, ${\cal A}$ might be an algorithm which fits a univariate linear or logistic regression which predicts $Y$ as a function of $\hat{Y}$.

\begin{algorithm}
\begin{spacing}{1.2}
\caption{A method for incorporating human expertise into algorithmic predictions}
\begin{algorithmic}[1]
\label{alg:meta_algo}
\State \textbf{Inputs:} Training data $\{x_i, y_i, \hat{y}_i\}_{i=1}^n$, a multicalibrated partition $\{S_k\}_{k \in K}$, a univariate regression algorithm ${\cal A}$, and a test observation $(x_{n+1}, \hat{y}_{n+1})$
\State \textbf{Output:} A prediction of the missing outcome $y_{n+1}$
\For {$k = 1$ to $K$}  \Comment{iterate over indistinguishable subsets}
    \State $Z_k \gets \{(\hat{y}_j, y_j) : x_j \in S_k\}$   \Comment{collect (expert prediction, outcome) pairs which lie in subset $k$}
    \State $\hat{g}_k \gets {\cal A}(Z_k)$  \Comment{regress outcomes on expert predictions, save resulting predictor $\hat{g}_k$}
\EndFor

\State $k^* \gets$ the index of the subset $S_k$ which contains $x_{n+1}$
\State \textbf{return} $\hat{g}_{k^*}(\hat{y}_{n+1})$ \Comment{return prediction corresponding to subset which contains $x_{n+1}$}

\end{algorithmic}
\end{spacing}
\end{algorithm}

The algorithm above simply learns a different predictor of $Y$ as a function of $\hat{Y}$ within each indistinguishable subset. As we show below, even simple instantiations of this approach can outperform the squared error achieved by \emph{any} $f \in \cF$. This approach can also be readily extended to more complicated human input --- for example, we might instead model a human's feedback as a high-dimensional vector instead of a point prediction $\hat{Y}$ --- and can be used to \emph{test} whether human judgment provides information that an algorithm cannot learn from the available training data. We turn to these results below.

\section{Technical results}
\label{sec: results}

In this section we present our main technical results. As outlined above, our method uses human predictions $\hY$ to predict the outcome $Y$ within subsets which are indistinguishable with respect to some function class $\cF$. In \Cref{thm: optimality of linear regression}, we show that a particularly simple instantiation this method --- a univariate linear regression of the outcome $Y$ on human predictions $\hY$ within each indistinguishable subset --- suffices to outperform the squared error of \emph{any} predictive algorithm $f \in \cF$. In \Cref{cor: high dimensional feedback}, we show that this result generalizes readily to arbitrary nonlinear predictors and/or cases where the expert provides richer, high-dimensional inputs (e.g., natural language feedback rather than a single point prediction $\hY \in [0, 1]$). In \Cref{thm: sufficiency of predictor implies bounded covariance}, we show that, in the important special case where $Y$ is a binary outcome, our method also provides an information-theoretic approach for proving that no subset of algorithmic predictors $F \subseteq \cF$ can obviate the need for human expertise (in a way we make precise below).

Proofs of \Cref{thm: optimality of linear regression}, \Cref{cor: high dimensional feedback} and \Cref{thm: sufficiency of predictor implies bounded covariance} are presented in \Cref{sec: proof_thm_1}, \Cref{sec: proof_cor_1} and \Cref{sec: proof_thm_2}, respectively. For clarity, results in this section are presented in terms of population quantities, and assume oracle access to a multicalibrated partition. We present corresponding generalization arguments and background on learning multicalibrated partitions in \Cref{sec:additional_tech} and \Cref{sec:learning partitions}, respectively. 

\begin{theorem}
\label{thm: optimality of linear regression}

Let $\{S_k\}_{k \in [K]}$ be an $\alpha$-multicalibrated partition with respect to a model class $\cF$ and target $Y$. Let the random variable $J(X) \in [K]$ be such that $J(X) = k$ iff $X \in S_k$. Define $\gamma^*, \beta^* \in \mathbb{R}^K$
as 
\begin{align}
\gamma^*, \beta^* & \in \argmin_{\gamma \in \mathbb{R}^K, \beta \in \mathbb{R}^K} \hspace{4pt} \E \left[ \left(Y - \gamma_{J(X)} + \beta_{J(X)} \hat{Y} \right)^2  \right].   \label{eq: least squares objective} 
\end{align}

Then, for any $f \in \mathcal{F}$ and $k \in [K]$,
\begin{align}
    \E_k\left[ \left(Y - \gamma^*_{k} - \beta^*_{k} \hY\right)^2 \right] + 4 \Cov_k(Y, \hY)^2 \le
    \E_k\left[ \left(Y - f(X) \right)^2 \right] + 2 \alpha.
\end{align}

\end{theorem}

That is, the squared error incurred by the univariate linear regression of $Y$ on $\hY$ within each indistinguishable subset outperforms that of any $f \in \cF$. This improvement is at least $4\Cov_k(Y, \hY)^2$, up to an additive approximation error $2\alpha$. We emphasize that $\cF$ is an arbitrary class, and may include complex, nonlinear predictors. Nonetheless, given a multicalibrated partition, a simple linear predictor can improve on the \emph{best} $f \in \cF$. Furthermore, this approach allows us to \emph{selectively} incorporate human feedback: whenever $\Cov_k(Y, \hY) = 0$, we recover a coefficient $\beta^*_k$ of $0$.\footnote{Recall that the population coefficient in a univariate linear regression of $Y$ on $\hY$ is $\frac{\Cov(Y, \hY)}{\Var(\hY)}$.}

\textbf{Nonlinear functions and high-dimensional feedback.} 
\Cref{thm: optimality of linear regression} corresponds to instantiating \Cref{alg:meta_algo} with a learning algorithm ${\cal A}$ that fits a univariate linear regression, but the same insight generalizes readily to other functional forms. For example, if the target $Y$ is a binary outcome, it might be desirable to instead fit a logistic regression. We provide a similar guarantee for generic nonlinear predictors via \Cref{cor: nonlinear predictors} in \Cref{sec:additional_tech}. Furthermore, while the results above assume that an expert provides a prediction $\hY \in [0, 1]$, the same insight extends to richer forms of feedback. For example, in a medical diagnosis task, a physician might produce free-form clinical notes which contain information that is not available in tabular electronic health records. Incorporating this kind of feedback requires a predictor which is better suited to high-dimensional inputs (e.g., a deep neural network), which motivates our following result.

\begin{corollary}\label{cor: high dimensional feedback}

Let $S$ be an $\alpha$-indistinguishable subset with respect to a model class $\cF$ and target $Y$. Let $H \in \cH$ denote expert feedback which takes values in some arbitrary domain (e.g,. $\reals^d$), and let $g: \cH \rightarrow [0, 1]$ be a function which satisfies the following approximate calibration condition for some $\eta \ge 0$ and for all $\beta, \gamma \in \reals$:

\begin{align}
    \E_S[(Y - g(H))^2] \le \E_S[(Y - \gamma - \beta g(H))^2] + \eta. \label{eq: approx calibration}
\end{align}

Then, for any $f \in \cF$, 

\begin{align}
     \E_S \left[ (Y - g(H))^2  \right] + 4 \Cov_S(Y, g(H))^2 \le
\E_S\left[ \left(Y - f(X) \right)^2 \right] + 2 \alpha + \eta.
\end{align}

\end{corollary}

To interpret this result, notice that \eqref{eq: approx calibration} requires only that the prediction $g(H)$ cannot be significantly improved by any linear post-processing function. For example, this condition is satisfied by any calibrated predictor $g(H)$.\footnote{A calibrated predictor is one where $\E_S[Y \mid g(H)] = g(H)$. This is a fairly weak condition; for example, it is satisfied by the constant predictor $g(H) \equiv \E_S[Y]$ \citep{well-calibrated-bayesian-1982, calibration-1988}.} Furthermore, any $g(H)$ which does not satisfy \eqref{eq: approx calibration} can be transformed by letting $\tilde{g}(H) = \min_{\gamma, \beta} \E[(Y - \gamma - \beta g(H))^2]$; i.e., by linearly regressing $Y$ on $g(H)$, in which case $\tilde{g}(H)$ satisfies \eqref{eq: approx calibration}. This result mirrors \Cref{thm: optimality of linear regression}: within each element of a multicalibrated partition, a predictor which depends only on human feedback $H$ can improve on the best $f \in \cF$. 

\textbf{Testing for informative experts.} While we have thus far focused on incorporating human judgment to improve predictions, a policymaker may also be interested in the related question of \emph{testing} whether human judgment provides information that cannot be expressed by any algorithmic predictor. For example, such a test might be valuable in informing whether to automate a given prediction task.

\Cref{thm: optimality of linear regression} suggests a heuristic for such a test: if the conditional covariance $\Cov_k(Y, \hY)$ is large, then we might expect that $\hY$ is somehow ``more informative'' than any $f \in \cF$ within $S_k$. While covariance only measures a certain form of \emph{linear} dependence between random variables, we now show that, in the special case of binary-valued algorithmic predictors, computing the covariance of $Y$ and $\hY$ within an indistinguishable subset serves as a stronger test for whether $\hY$ provides \emph{any} predictive information which cannot be expressed by the class $\cF$. We state this result as \Cref{thm: sufficiency of predictor implies bounded covariance} below.

\newpage

\begin{theorem}
\label{thm: sufficiency of predictor implies bounded covariance}
Let $\{S_k\}_{k \in [K]}$ be an $\alpha$-multicalibrated partition for a binary-valued model class $\cF^{\text{binary}}$ and target outcome $Y$. For all $k \in [K]$, let there be $\tilde{f}_k \in \cF$ such that $Y \indep \hY \mid \tilde{f}_k(X), X \in S_k$. Then, for all $k \in [K]$,
\begin{align}
    \left|\Cov_k(Y, \hY)\right| & \le \sqrt{\frac{\alpha}{2}}.
\end{align}
\end{theorem}

That is, if there exists a set of predictors $\{\tilde{f}_k\}_{k \in [K]}$ which ``explain" the signal provided by the human, then then the covariance of $Y$ and $\hY$ is bounded within each indistinguishable subset. The contrapositive implies that a sufficiently large value of $\Cov_k(Y, \hY)$ serves as a certificate for the property that \emph{no} subset of algorithmic predictors can fully explain the information that $\hY$ provides about $Y$. This result can be viewed as a finer-grained extension of the test proposed in \citet{auditing-expertise}. A proof of \Cref{thm: sufficiency of predictor implies bounded covariance} is provided in \Cref{sec: proof_thm_2}.

Taken together, our results demonstrate that algorithmic indistinguishability provides a principled way of reasoning about the complementary value of human judgment. Furthermore, this approach yields a concrete methodology for incorporating this expertise: we can simply use human feedback to predict $Y$ within subsets which are indistinguishable on the basis of $X$ alone. While our results in this section are stated in terms of population quantities, we provide a corresponding finite sample result in \Cref{sec:additional_tech}. Operationalizing these results depends critically on the ability to \emph{learn} multicalibrated partitions, e.g., via the boosting algorithm proposed in \citet{multicalibration-boosting-regression}. We provide additional detail on learning such partitions in \Cref{sec:learning partitions}.

\section{Experiments}
\label{sec:experiments}

In this section we return to the emergency room triage task described in \Cref{sec: introduction}. Recall that \Cref{fig: cond mcc} provided suggestive evidence that physicians have access to additional information which is unavailable to the Glasgow-Blatchford Score. We now apply the machinery developed in \Cref{sec: results} to investigate this more thoroughly; specifically, do physicians provide a predictive signal which cannot be extracted by any ``feasible" predictive algorithm? First, we provide additional background on the emergency room triage task.

\subsection{Risk stratification and triage for gastrointestinal bleeding}
\label{sec:GBS_background}
Acute gastrointestinal bleeding (AGIB) is a potentially serious condition for which 530,855 patients/year receive treatment in the United States alone \citep{PEERY2022621}. It is estimated that $32\%$ of patients with presumed bleeding from the lower gastrointestinal tract \citep{OAKLAND2017635} and $45\%$ of patients with presumed bleeding from the upper gastrointestinal tract \citep{Stanleyi6432} require urgent medical intervention; overall mortality rates for AGIB in the U.S. are estimated at around $3$ per $100,000$ \citep{PEERY2022621}. For patients who present with AGIB in the emergency room, the attending physician is tasked with deciding whether the bleeding is severe enough to warrant admission to the hospital. However, the specific etiology of AGIB is often difficult to determine from patient presentation alone, and gold standard diagnostic techniques --- an endoscopy for upper GI bleeding or a colonoscopy for lower GI bleeding --- are both invasive and costly, particularly when performed urgently in a hospital setting. As discussed in \Cref{sec: introduction}, the Glasgow-Blatchford Bleeding Score or GBS \citep{Blatchford2000-ce} is a standard screening metric used to algorithmically assess the risk that a patient with acute GI bleeding will require red blood cell transfusion, intervention to stop bleeding, or die within 30 days.\footnote{US and international guidelines use the Glasgow-Blatchford score as the preferred risk score for assessing patients with upper gastrointestinal bleeding \citep{Laine2021-oe,BARKUN2019}; it has been also validated in patients with acute lower gastrointestinal bleeding \citep{URRAHMAN2018}. Other risk scores tailored to bleeding in the lower gastrointestinal tract have been proposed in the literature, but these are less commonly used in practice. In this work, we interpret the GBS as a measure of risk for patients who present with either upper or lower GI bleeding, and refer interested readers to \citet{Almaghrabi2022-sp} for additional background.}

\textbf{Construction of the Glasgow-Blatchford Score.} The Glasgow-Blatchford Score is a function of the following nine patient characteristics, which are converted to discrete values before being provided as input: blood urea nitrogen (BUN), hemoglobin (HGB), systolic blood pressure (SBP), pulse, cardiac failure, hepatic disease, melena, syncope and biological sex.
Scores are integers ranging from $0$ to $23$, with higher scores indicating a higher risk that a patient will require subsequent intervention. US and international guidelines suggest that patients with a score of $0$ or $1$ can be safely discharged from the emergency department  \citep{BARKUN2019, Laine2021-oe}, with further investigation to be performed outside the hospital. For additional details on the construction of the GBS, we refer to \citet{Blatchford2000-ce}.

\textbf{Defining features, predictions and outcomes.} We consider 3617 patients who presented with AGIB at one of three hospitals in a large academic health system between 2014 and 2018. Consistent with the goals of triage for patients with AGIB, we record an \emph{adverse outcome} if a patient (1) requires some form of urgent intervention to stop bleeding (endoscopic, interventional radiologic, or surgical; excluding patients who only undergo a diagnostic endoscopy or colonoscopy) while in the hospital (2) dies within 30 days of their emergency room visit or (3) is initially discharged but later readmitted within $30$ days.\footnote{This threshold is consistent with the definition used in the Centers for Medicare and Medicaid Services Hospital Readmission Reduction Program, which seeks to incentivize healthcare providers to avoid discharging patients who will be readmitted within 30 days \citep{doi:10.1056/CAT.18.0194}} As is typical of large urban hospitals in the United States, staffing protocols at this health system dictate a separation of responsibilities between emergency room physicians and other specialists. In particular, while emergency room physicians make an initial decision whether to hospitalize a patient, it is typically a gastrointestinal specialist who subsequently decides whether a patient admitted with AGIB requires some form of urgent hemostatic intervention. This feature, along with our ability to observe whether discharged patients are subsequently readmitted, substantially mitigates the selective labels issue that might otherwise occur in this setting. Thus, consistent with clinical and regulatory guidelines --- to avoid hospitalizing patients who do not require urgent intervention \citep{Stanley2009-gq}, and to avoid discharging patients who are likely to be readmitted within 30 days \citep{doi:10.1056/CAT.18.0194} --- we interpret the emergency room physician's decision to admit or discharge a patient as a \emph{prediction} that one of these adverse outcomes will occur. We thus instantiate our model by letting $X_i \in \mathbb{N}^9$ be the nine discrete patient characteristics from which the Glasgow-Blatchford Score is computed. We further let $\hat{Y}_i \in \{0, 1\}$ indicate whether that patient was initially hospitalized, and $Y_i \in \{0, 1\}$ indicate whether that patient suffered one of the adverse outcomes defined above. 

\textbf{Assessing the accuracy of physician decisions.} We first summarize the performance of the emergency room physicians' hospitalization decisions, and compare them with the performance of a simple rule which would instead admit every patient with a GBS above a certain threshold and discharge the remainder in \Cref{tab:physician and gbs perf}. We consider thresholds of $0$ and $1$ -- the generally accepted range for low risk patients (\citet{Laine2021-oe, BARKUN2019}) -- as well as less conservative thresholds of $2$ and $7$ (the latter of which we find maximizes overall accuracy). For additional context, we also provide the total fraction of patients admitted under each decision rule.

\begin{table}[!htbp]
\centering
\captionsetup{justification=centering}
\begin{tabular}{lllll}
  \toprule Decision Rule & Fraction Hospitalized & Accuracy & Sensitivity & Specificity \\ 
  \midrule Physician Discretion & 0.86 ± 0.02 & 0.55 ± 0.02 & 0.99 ± 0.00 & 0.24 ± 0.02 \\ 
  Admit GBS $>$ 0 & 0.88 ± 0.02 & 0.53 ± 0.02 & 0.99 ± 0.00 & 0.19 ± 0.02 \\ 
  Admit GBS $>$ 1 & 0.80 ± 0.02 & 0.60 ± 0.02 & 0.98 ± 0.00 & 0.33 ± 0.02 \\ 
  Admit GBS $>$ 2 & 0.73 ± 0.02 & 0.66 ± 0.02 & 0.97 ± 0.00 & 0.43 ± 0.02 \\ 
  Admit GBS $>$ 7 & 0.40 ± 0.02 & 0.79 ± 0.02 & 0.73 ± 0.02 & 0.84 ± 0.02 \\ 
   \bottomrule \end{tabular}
\caption{
Comparing the accuracy of physician hospitalization decisions to thresholding the GBS. A decision is correct if it hospitalizes a patient who suffers an adverse outcome (as defined above) or discharges a patient who does not. Results are reported to $\pm 2$ standard errors. 
} 
\label{tab:physician and gbs perf}
\end{table}

Unsurprisingly, we find that the physicians calibrate their decisions to maximize sensitivity (minimize false negatives) at the expense of admitting a significant fraction of patients who, in retrospect, could have been discharged immediately. Indeed we find that although $86\%$ of patients are initially hospitalized, only $\approx 42\%$ actually require a specific hospital-based intervention or otherwise suffer an adverse outcome which would justify hospitalization (i.e., $y_i = 1$). Consistent with \citet{Blatchford2000-ce} and \citet{Chatten2018-co}, we also find that thresholding the GBS in the range of $[0, 2]$ achieves a sensitivity of close to $100\%$. We further can see that using one of these thresholds achieves overall accuracy (driven by improved specificity) which is substantially better than physician discretion.

\subsection{Indistinguishability with respect to an infinite class}\label{sec: infinite-indistinguishability} \Cref{tab:physician and gbs perf} demonstrates that the Glasgow-Blatchford Score is highly competitive with physicians, at least \emph{on average} over the patient population. However, in \Cref{sec: introduction}, we saw suggestive evidence that physicians incorporate additional information which is not encoded in the inputs to the GBS when making triage decisions. We now look to reconcile these findings by investigating whether physicians can selectively distinguish patients who are algorithmically indistinguishable, and thus improve predictive performance. 

In particular, to illustrate the flexibility of our approach, we consider indistinguishability with respect to an infinitely large but nonetheless simple class of shallow (depth $\le 3$) regression trees. We denote this class by $\cF^{\text{RT3}}$. We define our input space as the nine features which serve as input to the GBS (as described in \Cref{sec:GBS_background}), as well as a tenth feature which is the Glasgow-Blatchford Score itself. Intuitively, we will iteratively construct a partition in which the patients in each subset cannot be further distinguished by any shallow regression tree. Then, within each subset, we will test whether physician judgment adds additional predictive signal to distinguish patients who should be hospitalized from those who can be safely discharged. Importantly, this partition is also indistinguishable with respect to the GBS, as no shallow regression tree --- which, in particular, can directly split on the value of the GBS --- can further distinguish patients within each indistinguishable subset. Thus, indistinguishability with respect to $\cF^{\text{RT3}}$ implies indistinguishability with respect to the GBS alone (and the latter is a substantially weaker condition than the former).

To learn this partition, we apply the boosting algorithm proposed in \citet{multicalibration-boosting-regression} to construct a predictor $h: \cX \rightarrow [0, 1]$ such that no $f \in \cF^{\text{RT}3}$ can substantially improve on the squared error of $h$ within any of its approximate level sets $\{ x \mid h(x) \in [0, .1)\}, \{x \mid h(x) \in (.1, .2] \dots \{x \mid h(x) \in [.9, 1)\}, \{x \mid h(x) = 1\}$.\footnote{We discuss the connection between this condition and multicalibration in \Cref{sec:learning partitions}. We cannot necessarily verify whether the multicalibration condition is satisfied empirically; however, the theory provides guidance for choosing subsets, within which we can directly test the conditional performance physicians. We restrict the algorithm to require at least $50$ patients within each level set for an update to occur; for additional details, including the recommended hyperparameters which we adopt, we refer to \cite{multicalibration-boosting-regression}.} We learn the predictor $h$ on the the training set containing $2893$ patients, and plot the Matthew's Correlation Coefficient between physician decisions and adverse outcomes on the remaining $724$ patients within each level set in \Cref{fig: mc-distinguishability} below.

\begin{figure}[htpb!]
\includegraphics[scale=.50]{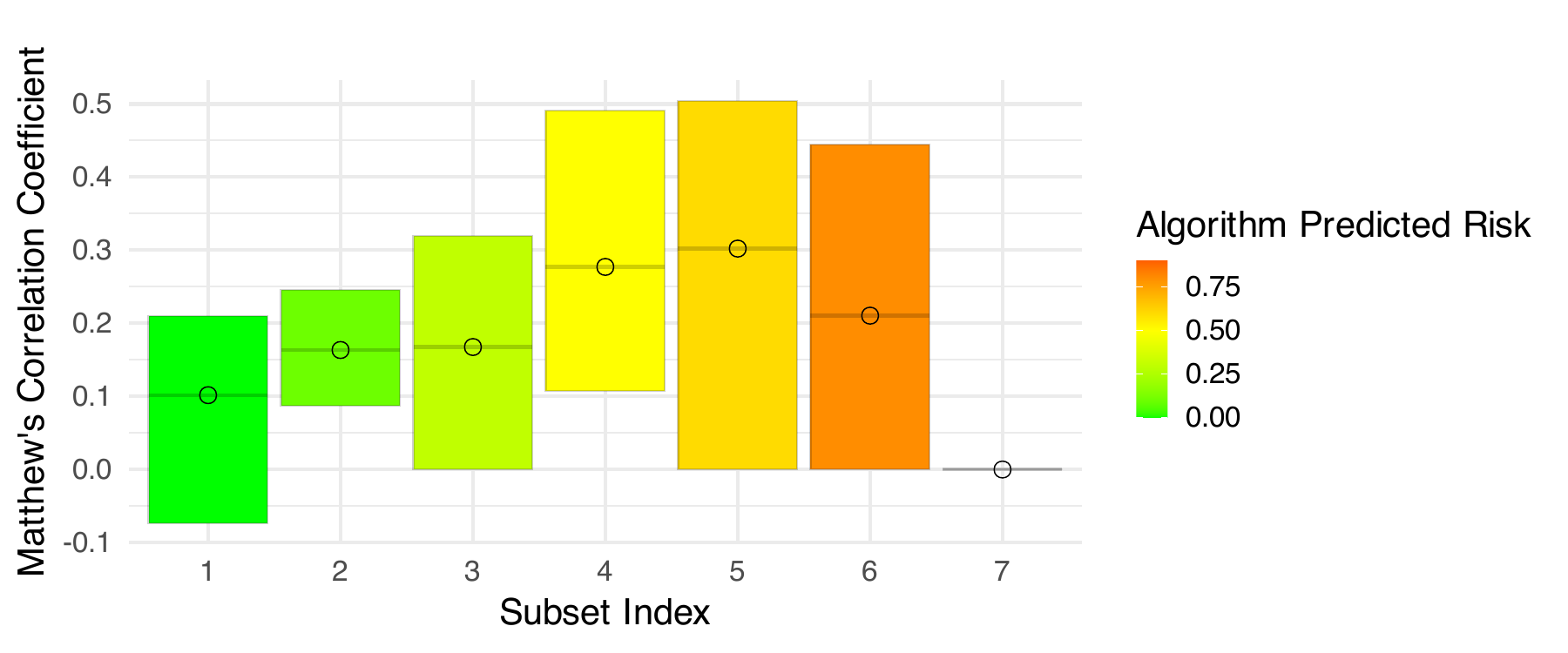}
\captionsetup{justification=centering}
    \centering
    \caption{Physician performance within the level sets of a predictor which is multicalibrated with respect to $\cF^{\text{RT}3}$. Point estimates are reported with $95\%$ Bonferroni corrected bootstrap confidence intervals (estimated from $1000$ bootstrap replicates).}
    \label{fig: mc-distinguishability}
\end{figure}

\Cref{fig: mc-distinguishability} indicates that physicians appear to add a predictive signal which cannot be extracted from the data by any shallow regression tree, as their decisions remain meaningfully correlated with adverse event risk within the level sets of a predictor which is multicalibrated over $\cF^{\text{RT}3}$. It is again instructive to directly compare \Cref{fig: mc-distinguishability} to \Cref{fig: cond mcc} and \Cref{fig: by_gbs}, as the analysis in this section provides a natural way of interpolating between these two extremes. On one hand, examining physician performance within observationally indistinguishable subsets (\Cref{fig: cond mcc}) is guaranteed to directly test whether physicians provide signal which is not present in the \emph{data}, but partitions the input space too finely to be statistically tractable. On the other hand, examining physician performance within the level sets of the GBS (\Cref{fig: by_gbs}) indicated that physicians provide signal that is not expressed by the GBS, but does not tell us whether a \emph{different} predictive algorithm might obviate the need for physician expertise. \Cref{fig: mc-distinguishability} demonstrates the result of the natural compromise suggested by our framework, as we are able to test whether physicians add signal that cannot be extracted by \emph{any} predictor in a rich class of predictive algorithms while only testing physician performance within a modest number of indistinguishable subsets (here, $7$). Given additional training data, we could naturally extend this approach to successively richer function classes, thus approaching the strongest claim (sketched in \Cref{fig: cond mcc}) that physicians provide signal which cannot be extracted by \emph{any} algorithmic predictor.

\textbf{Physician judgment does not improve mean squared error.} Given the multicalibrated partition in \Cref{fig: mc-distinguishability}, we can directly apply \Cref{thm: optimality of linear regression} to test whether physician judgment can improve mean squared error over the entire patient population. In particular, we first fit a univariate linear regression of the binary outcome $Y$ on the physician decision $\hat{Y}$ within each of the $7$ indistinguishable subsets, using the same subsample of $2893$ randomly selected patients. We then evaluate these predictions on the remaining $724$ patients, where we find that the mean squared error achieved by incorporating physician judgment $(0.1362; 95\%\ \text{CI}: 0.1211 - 0.1541)$ fails to significantly outperform the mean squared error achieved by the predictor which simply predicts the mean value of $Y$ (computed on the training set) within each indistinguishable subset $(0.1442; 95\%\ \text{CI}: 0.1278 - 0.1609)$.\footnote{Intuitively, the mean value of $Y$ within each subset can be interpreted as the best predictor which does not incorporate physician judgment (because patients cannot be further distinguished by any $f \in \cF^{\text{RT}3}$). For additional discussion, see the proof of \Cref{thm: optimality of linear regression} in \Cref{sec: proof_thm_1}. $95\%$ confidence intervals are calculated over $1000$ bootstrap samples.} However, while incorporating physician judgment does not appear to improve mean squared error \emph{on average} over the patient population, \Cref{fig: mc-distinguishability} suggests that \emph{selectively} incorporating physician judgment may still improve performance. Furthermore, as discussed in \Cref{sec:GBS_background}, physicians are ultimately responsible for making \emph{decisions} for each patient, and likely have preferences over outcomes which are substantially richer than simply minimizing mean squared error (e.g., minimizing false positives). In the following section, we develop a simple decision-theoretic framework which demonstrates how algorithmic indistinguishability provides a basis for incorporating such preferences.

\subsection{Leveraging physician judgment to improve decision making}

As discussed above, the results in \Cref{sec: case study} and \Cref{sec: infinite-indistinguishability} suggest that physician decisions provide \emph{information} that algorithmic predictors cannot learn from the training data. It is thus natural to ask whether this heterogeneity allows us to design better \emph{decision rules} (hereafter, ``policies") which leverage the complementary strengths of algorithms and physicians. Specifically, a natural class of policies to consider is those which make one of three decisions for each indistinguishable subset: hospitalize every patient, discharge every patient, or defer to the physician's decision for every patient. Intuitively, because an algorithmic predictor cannot distinguish individuals \emph{within} an indistinguishable subset, it is natural to constrain the policy space to those which make the same decision on all patients which are not deferred to the physician. This yields only $3^7 = 2187$ possible policy choices, as we make one of $3$ decisions within each of the $7$ subsets in \Cref{fig: mc-distinguishability}. In contrast, there are $669$ unique patients in our training set and, in principle, more than $5000$ unique combinations of the $9$ discrete-valued patient characteristics.

It is not immediately clear how to rank these policies, as the choice naturally depends on the policy makers' preferences over different measures of accuracy (e.g., false positive and false negative rates), the fraction of decisions which are automated (e.g., due to constraints on physicians' bandwidth) and other unobserved factors. Although modeling the decision maker's utility function is beyond the scope of this work, we provide an overview of each possible policy evaluated on the test set in \Cref{fig: policies} below. We exclude policies which are Pareto dominated with respect to the three measures we consider i.e., those which simultaneously have a weakly smaller true positive rate, weakly larger false positive rate and weakly smaller fraction of cases automated, where at least one inequality is strict. We emphasize that this is only intended to give a sample of the attractive policy choices afforded by our approach; it is possible that policymakers indeed have richer preferences (e.g., that include equity or fairness concerns) which would lead to a different notion of pareto dominance.

\begin{figure}[h!]
\includegraphics[scale=.75]{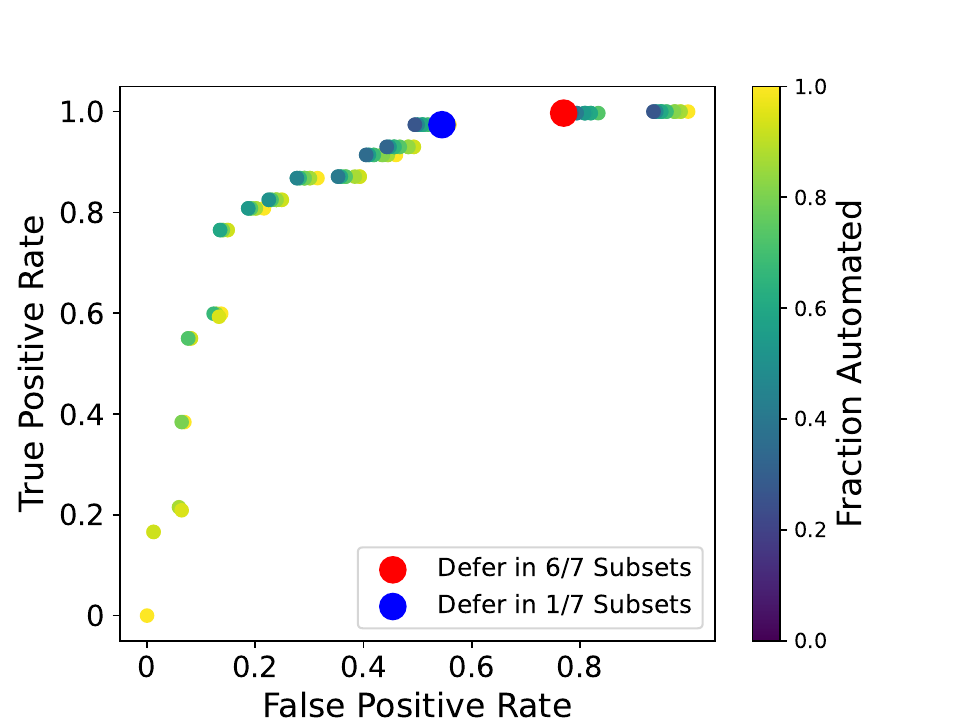}
\captionsetup{justification=centering}
    \centering
    \caption{The performance of policies which independently choose whether to hospitalize, discharge, or defer to the physician within each indistinguishable subset. The red policy defers to the physician within all but one subset; it achieves a true positive rate of $99.7\%$, a false positive rate of $77.0\%$, and automates $7.6\%$ of decisions. The blue policy only defers to the physician within one subset, achieving a true positive rate of $97.4\%$ and a false positive rate of $53.6\%$ while automating $86\%$ of decisions.}
    \label{fig: policies}
\end{figure}

 As \Cref{fig: policies} demonstrates, policies of this form yield a diverse array of tradeoffs between false positive rates, true positive rates and the fraction of automated decisions. For example, the policy highlighted in red defers to the physician within all but the highest risk subset, where the algorithm predicts that patients have a $\ge 90\%$ adverse event risk. Within this subset, the red policy hospitalizes every patient, thus automating the hospitalization decision for $7.6\%$ of the highest risk patients. This policy \emph{exactly} matches the true positive and false positive rate of the physician, as the physician also chooses to hospitalize every patient in the highest risk subset. Importantly, US and international clinical guidelines  \citep{BARKUN2019, Laine2021-oe} do not make  specific recommendations regarding the use of algorithmic decision rules for the highest risk patients; however, our results suggest that algorithms can help enhance patient safety by ensuring that all high-risk patients receive immediate escalation of care. This suggests a complementary role for algorithms in making triage and treatment decisions for the \emph{highest risk} patients. 
 
 In contrast, the blue policy achieves a slightly lower true positive rate of $97.4\%$, but compensates for this loss with a large decrease in the false positive rate ($53.6\%$ vs the physician's $77.0\%$) while automating $86\%$ of decisions. Under this policy, the decision to discharge or hospitalize patients is automated in every subset except for subset $4$ (where, as \Cref{fig: mc-distinguishability} suggests, physician judgment provides the most additional value). We caution that this latter policy may be inappropriate for real-world triage decisions, as both US and international guidelines indicate that the consensus is that algorithmic decision rules (or algorithmic recommendations) should achieve a true positive rate of at least $99\%$ \citep{BARKUN2019, Laine2021-oe}. Nonetheless, this serves to illustrate the manner in which indistinguishability can provide decision makers with an array of potentially attractive policies. Furthermore, additional data might allow for finer-grained stratification of patients by learning a multicalibrated partition over a richer function class, which in turn could yield a richer space of possible decision rules.

\section{Discussion}
\label{sec:discussion}
 In this work we propose a framework for enabling human-AI collaboration in prediction and decision tasks. Under this framework, we develop a set of methods for testing whether experts provide a predictive signal which cannot be replicated by an algorithmic predictor, and further show how to selectively incorporate human feedback to improve algorithmic predictions.  In \Cref{sec:experiments}, we show how algorithmic indistinguishability can induce a natural set of decision rules which govern when an algorithm should defer to a human expert. Beyond these methodological contributions, we argue that our framing clarifies \emph{when} and \emph{why} human judgment can improve algorithmic performance. In particular, a key theme in our work is that even if humans do not consistently outperform algorithms on average, \emph{selectively} incorporating human judgment can often improve both predictions and decision-making.

A key limitation of our work is a somewhat narrow focus on minimizing a well-defined loss function over a well-defined (and stationary) distribution. This fails to capture decision makers with richer, multidimensional preferences that may be difficult to model (e.g., fairness, robustness or simplicity), and does not extend to settings in which predictions \emph{influence} future outcomes (see the discussion of performative prediction in \Cref{sec:related}). However, we view indistinguishability as a powerful primitive for modeling these more complex scenarios; for example, a decision maker might impose additional preferences --- like a desire for some notion of fairness --- to distinguish inputs which are otherwise indistinguishable with respect to the ``primary" outcome of interest.

At a technical level, our results rely on the ability to efficiently \emph{learn} multicalibrated partitions. While we give conditions under which this is feasible in \Cref{sec:learning partitions} and a finite sample analysis of our main results in \Cref{sec:additional_tech}, finding such partitions can be challenging for rich function classes. Furthermore, as \Cref{fig:mc_stylized_example} demonstrates, multicalibrated partitions are not always unique, and our work does not offer guidance for choosing among multiple possibilities. However, as discussed in \Cref{sec:experiments}, algorithms for learning multicalibrated partitions still offer useful heuristics, as the key question --- does incorporating human feedback improve algorithmic predictions within a given subset of the input space --- remains empirically testable even when the assumptions underpinning our theory are violated. 

Finally, we caution that even in contexts which fit neatly into our framework, human decision makers can be critical for ensuring interpretability and accountability. Thus, although our approach can provide guidance for choosing the appropriate level of automation, it does not address practical or ethical concerns which may arise. Despite these limitations, we argue that indistinguishability helps to clarify the role of human expertise in algorithmic decision making, and this framing in turn provides fundamental conceptual and methodological insights for enabling effective human-AI collaboration.

\section*{Acknowledgements}
Rohan Alur, Devavrat Shah and Manish Raghavan are supported in part by a Stephen A. Schwarzman College of Computing Seed Grant. Devavrat Shah is supported in part by an NSF FODSI Award (2022448). Dennis Shung is supported in part by an NIH NIDDK Award (DK125718).

\bibliographystyle{apalike}
\bibliography{references}

\begin{thebibliography}{}

\bibitem[Agarwal et~al., 2023]{human-expertise-ai-2023}
Agarwal, N., Moehring, A., Rajpurkar, P., and Salz, T. (2023).
\newblock Combining human expertise with artificial intelligence: Experimental evidence from radiology.
\newblock Technical report, Cambridge, MA.

\bibitem[Agrawal et~al., 2018]{prediction-vs-judgment-2018}
Agrawal, A., Gans, J., and Goldfarb, A. (2018).
\newblock Exploring the impact of artificial intelligence: Prediction versus judgment.
\newblock Technical report, National Bureau of Economic Research, Cambridge, MA.

\bibitem[Almaghrabi et~al., 2022]{Almaghrabi2022-sp}
Almaghrabi, M., Gandhi, M., Guizzetti, L., Iansavichene, A., Yan, B., Wilson, A., Oakland, K., Jairath, V., and Sey, M. (2022).
\newblock Comparison of risk scores for lower gastrointestinal bleeding: A systematic review and meta-analysis.
\newblock {\em JAMA Netw. Open}, 5(5):e2214253.

\bibitem[Alur et~al., 2023]{auditing-expertise}
Alur, R., Laine, L., Li, D.~K., Raghavan, M., Shah, D., and Shung, D.~L. (2023).
\newblock Auditing for human expertise.
\newblock In {\em Advances in Neural Information Processing Systems}, volume~36.

\bibitem[Arnold et~al., 2020]{discrimination-bail-decisions-2020}
Arnold, D., Dobbie, W., and Hull, P. (2020).
\newblock Measuring racial discrimination in bail decisions.
\newblock Technical report, National Bureau of Economic Research, Cambridge, MA.

\bibitem[Bansal et~al., 2020]{accuracy-teamwork-2020}
Bansal, G., Nushi, B., Kamar, E., Horvitz, E., and Weld, D.~S. (2020).
\newblock Is the most accurate ai the best teammate? optimizing ai for teamwork.

\bibitem[Barkun et~al., 2019]{BARKUN2019}
Barkun, A.~N., Almadi, M., Kuipers, E.~J., Laine, L., Sung, J., Tse, F., Leontiadis, G.~I., Abraham, N.~S., Calvet, X., Francis K.L.~Chan, J.~D., Enns, R., Gralnek, I.~M., Jairath, V., Jensen, D., Lau, J., Lip, G.~Y., Loffroy, R., Maluf-Filho, F., Meltzer, A.~C., Reddy, N., Saltzman, J.~R., Marshall, J.~K., and Bardou, M. (2019).
\newblock Management of nonvariceal upper gastrointestinal bleeding: Guideline recommendations from the international consensus group.
\newblock {\em Annals of Internal Medicine}, 171(11):805--822.

\bibitem[Bastani et~al., 2021]{improving-decisions-2021}
Bastani, H., Bastani, O., and Sinchaisri, W.~P. (2021).
\newblock Improving human decision-making with machine learning.

\bibitem[Beede et~al., 2020]{human-centered-eval-2020}
Beede, E., Baylor, E.~E., Hersch, F., Iurchenko, A., Wilcox, L., Ruamviboonsuk, P., and Vardoulakis, L.~M. (2020).
\newblock A human-centered evaluation of a deep learning system deployed in clinics for the detection of diabetic retinopathy.
\newblock {\em Proceedings of the 2020 CHI Conference on Human Factors in Computing Systems}.

\bibitem[Benz and Rodriguez, 2023]{human-aligned-calibration-2023}
Benz, N. L.~C. and Rodriguez, M.~G. (2023).
\newblock Human-aligned calibration for ai-assisted decision making.

\bibitem[Blatchford et~al., 2000]{Blatchford2000-ce}
Blatchford, O., Murray, W.~R., and Blatchford, M. (2000).
\newblock A risk score to predict need for treatment for upper-gastrointestinal haemorrhage.
\newblock {\em Lancet}, 356(9238):1318--1321.

\bibitem[Brown et~al., 2020]{Brown2020PerformativePI}
Brown, G., Hod, S., and Kalemaj, I. (2020).
\newblock Performative prediction in a stateful world.
\newblock {\em ArXiv}, abs/2011.03885.

\bibitem[Camerer and Johnson, 1991]{process-performance-paradox-1991}
Camerer, C. and Johnson, E. (1991).
\newblock The process-performance paradox in expert judgment: How can experts know so much and predict so badly?
\newblock In Ericsson, A. and Smith, J., editors, {\em Toward a General Theory of Expertise: Prospects and Limits}. Cambridge University Press.

\bibitem[Chatten et~al., 2018]{Chatten2018-co}
Chatten, K., Purssell, H., Banerjee, A.~K., Soteriadou, S., and Ang, Y. (2018).
\newblock Glasgow blatchford score and risk stratifications in acute upper gastrointestinal bleed: can we extend this to 2 for urgent outpatient management?
\newblock {\em Clin. Med.}, 18(2):118--122.

\bibitem[Chicco and Jurman, 2020]{advantages-mcc-2020}
Chicco, D. and Jurman, G. (2020).
\newblock The advantages of the matthews correlation coefficient ({MCC}) over {F1} score and accuracy in binary classification evaluation.
\newblock {\em BMC Genomics}, 21(1):6.

\bibitem[Cook, 2009]{Cook2009}
Cook, C. (2009).
\newblock Is clinical gestalt good enough?
\newblock {\em Journal of Manual and Manipulative Therapy}, 17(1):6--7.

\bibitem[Cowgill, 2018]{bias-productivity-2018}
Cowgill, B. (2018).
\newblock Bias and productivity in humans and algorithms: Theory and evidence from resume screening.

\bibitem[Cowgill and Stevenson, 2020]{algorithmic-social-eng-2020}
Cowgill, B. and Stevenson, M.~T. (2020).
\newblock Algorithmic social engineering.
\newblock {\em AEA Pap. Proc.}, 110:96--100.

\bibitem[Currie and MacLeod, 2017]{diagnosing-expertise-2017}
Currie, J. and MacLeod, W.~B. (2017).
\newblock Diagnosing expertise: Human capital, decision making, and performance among physicians.
\newblock {\em J. Labor Econ.}, 35(1):1--43.

\bibitem[Dawes, 1971]{graduate-admissions-1971}
Dawes, R.~M. (1971).
\newblock A case study of graduate admissions: Application of three principles of human decision making.
\newblock {\em Am. Psychol.}, 26(2):180--188.

\bibitem[Dawes et~al., 1989]{clinical-v-actuarial-1989}
Dawes, R.~M., Faust, D., and Meehl, P.~E. (1989).
\newblock Clinical versus actuarial judgment.
\newblock {\em Science}, 243(4899):1668--1674.

\bibitem[Dawid, 1982]{well-calibrated-bayesian-1982}
Dawid, A.~P. (1982).
\newblock The well-calibrated bayesian.
\newblock {\em Journal of the American Statistical Association}, 77:605--610.

\bibitem[De et~al., 2020]{classification-under-assistance-2020}
De, A., Okati, N., Zarezade, A., and Gomez-Rodriguez, M. (2020).
\newblock Classification under human assistance.

\bibitem[De-Arteaga et~al., 2020]{case-for-hil-2020}
De-Arteaga, M., Fogliato, R., and Chouldechova, A. (2020).
\newblock A case for humans-in-the-loop: Decisions in the presence of erroneous algorithmic scores.

\bibitem[Dietvorst et~al., 2018]{overcoming-aversion-2018}
Dietvorst, B., Simmons, J., and Massey, C. (2018).
\newblock Overcoming algorithm aversion: People will use imperfect algorithms if they can (even slightly) modify them.
\newblock {\em Management Science}, 64:1155--1170.

\bibitem[Donahue et~al., 2022]{hai-complementarity-2022}
Donahue, K., Chouldechova, A., and Kenthapadi, K. (2022).
\newblock Human-algorithm collaboration: Achieving complementarity and avoiding unfairness.

\bibitem[Dwork et~al., 2020]{outcome-indistinguishability}
Dwork, C., Kim, M.~P., Reingold, O., Rothblum, G.~N., and Yona, G. (2020).
\newblock Outcome indistinguishability.

\bibitem[Foster and Vohra, 1998]{calibration-1988}
Foster, D.~P. and Vohra, R.~V. (1998).
\newblock Asymptotic calibration.
\newblock {\em Biometrika}, 85(2):379--390.

\bibitem[Globus-Harris et~al., 2023]{multicalibration-boosting-regression}
Globus-Harris, I., Harrison, D., Kearns, M., Roth, A., and Sorrell, J. (2023).
\newblock Multicalibration as boosting for regression.

\bibitem[Gopalan et~al., 2022]{omnipredictors}
Gopalan, P., Kalai, A.~T., Reingold, O., Sharan, V., and Wieder, U. (2022).
\newblock Omnipredictors.
\newblock In {\em Innovations in Theoretical Computer Science (ITCS'2022)}.

\bibitem[Gopalan et~al., 2023]{omniprediction-via-multicalibration}
Gopalan, P., Kim, M.~P., and Reingold, O. (2023).
\newblock Swap agnostic learning, or characterizing omniprediction via multicalibration.

\bibitem[Grove et~al., 2000]{clinical-v-mechanical-2000}
Grove, W.~M., Zald, D.~H., Lebow, B.~S., Snitz, B.~E., and Nelson, C. (2000).
\newblock {{C}linical versus mechanical prediction: a meta-analysis}.
\newblock {\em Psychol Assess}, 12(1):19--30.

\bibitem[Hardt and Mendler-D{\"u}nner, 2023]{Hardt2023PerformativePP}
Hardt, M. and Mendler-D{\"u}nner, C. (2023).
\newblock Performative prediction: Past and future.
\newblock {\em ArXiv}, abs/2310.16608.

\bibitem[H{\'e}bert-Johnson et~al., 2018]{multicalibration-2017}
H{\'e}bert-Johnson, U., Kim, M., Reingold, O., and Rothblum, G. (2018).
\newblock Multicalibration: Calibration for the (computationally-identifiable) masses.
\newblock In {\em International Conference on Machine Learning}, pages 1939--1948. PMLR.

\bibitem[Jagadeesan et~al., 2022]{Jagadeesan2022RegretMW}
Jagadeesan, M., Zrnic, T., and Mendler-D{\"u}nner, C. (2022).
\newblock Regret minimization with performative feedback.
\newblock In {\em International Conference on Machine Learning}.

\bibitem[Kabrhel et~al., 2005]{Kabrhel2005}
Kabrhel, C., Camargo, C. A.~J., and Goldhaber, S.~Z. (2005).
\newblock Clinical gestalt and the diagnosis of pulmonary embolism: Does experience matter?
\newblock {\em Chest}, 127(5):1627--1630.

\bibitem[Keswani et~al., 2021]{multiple-experts-2021}
Keswani, V., Lease, M., and Kenthapadi, K. (2021).
\newblock Towards unbiased and accurate deferral to multiple experts.
\newblock {\em Proceedings of the 2021 AAAI/ACM Conference on AI, Ethics, and Society}.

\bibitem[Keswani et~al., 2022]{closed-deferral-pipelines-2022}
Keswani, V., Lease, M., and Kenthapadi, K. (2022).
\newblock Designing closed human-in-the-loop deferral pipelines.

\bibitem[Kim and Perdomo, 2022]{Kim2022MakingDU}
Kim, M.~P. and Perdomo, J.~C. (2022).
\newblock Making decisions under outcome performativity.
\newblock {\em ArXiv}, abs/2210.01745.

\bibitem[Klaes and Sent, 2005]{bounded-rationality-history-2005}
Klaes, M. and Sent, E.-M. (2005).
\newblock A conceptual history of the emergence of bounded rationality.
\newblock {\em Hist. Polit. Econ.}, 37(1):27--59.

\bibitem[Kleinberg et~al., 2017]{human-decisions-machine-predictions-2017}
Kleinberg, J., Lakkaraju, H., Leskovec, J., Ludwig, J., and Mullainathan, S. (2017).
\newblock Human decisions and machine predictions.

\bibitem[Kleinberg et~al., 2015]{prediction-policy-2015}
Kleinberg, J., Ludwig, J., Mullainathan, S., and Obermeyer, Z. (2015).
\newblock Prediction policy problems.
\newblock {\em American Economic Review}, 105(5):491--95.

\bibitem[Kuncel et~al., 2013a]{mechanical-vs-clinical-2013}
Kuncel, N.~R., Klieger, D.~M., Connelly, B.~S., and Ones, D.~S. (2013a).
\newblock {{M}echanical versus clinical data combination in selection and admissions decisions: a meta-analysis}.
\newblock {\em J Appl Psychol}, 98(6):1060--1072.

\bibitem[Kuncel et~al., 2013b]{mechanical-v-clinical}
Kuncel, N.~R., Klieger, D.~M., Connelly, B.~S., and Ones, D.~S. (2013b).
\newblock {{M}echanical versus clinical data combination in selection and admissions decisions: a meta-analysis}.
\newblock {\em J Appl Psychol}, 98(6):1060--1072.

\bibitem[Laine et~al., 2021]{Laine2021-oe}
Laine, L., Barkun, A.~N., Saltzman, J.~R., Martel, M., and Leontiadis, G.~I. (2021).
\newblock {ACG} clinical guideline: Upper gastrointestinal and ulcer bleeding.
\newblock {\em Am. J. Gastroenterol.}, 116(5):899--917.

\bibitem[Lakkaraju et~al., 2017]{selective-labels-2017}
Lakkaraju, H., Kleinberg, J., Leskovec, J., Ludwig, J., and Mullainathan, S. (2017).
\newblock The selective labels problem: Evaluating algorithmic predictions in the presence of unobservables.
\newblock In {\em Proceedings of the 23rd ACM SIGKDD International Conference on Knowledge Discovery and Data Mining}, KDD '17, page 275–284, New York, NY, USA. Association for Computing Machinery.

\bibitem[Leit{\~a}o et~al., 2022]{beyond-learning-to-defer-2022}
Leit{\~a}o, D., Saleiro, P., Figueiredo, M. A.~T., and Bizarro, P. (2022).
\newblock Human-ai collaboration in decision-making: Beyond learning to defer.
\newblock {\em ArXiv}, abs/2206.13202.

\bibitem[Madras et~al., 2018]{learning-to-defer-2018}
Madras, D., Pitassi, T., and Zemel, R.~S. (2018).
\newblock Predict responsibly: Improving fairness and accuracy by learning to defer.
\newblock In {\em Advances in Neural Information Processing Systems 31}, pages 6150--6160.

\bibitem[Marwaha et~al., 2022]{Marwaha2022}
Marwaha, J.~S., Landman, A.~B., Brat, G.~A., Dunn, T., and Gordon, W.~J. (2022).
\newblock Deploying digital health tools within large, complex health systems: key considerations for adoption and implementation.
\newblock {\em npj Digital Medicine}, 5:13.

\bibitem[Meehl, 1954]{Meehl1954ClinicalVS}
Meehl, P.~E. (1954).
\newblock Clinical versus statistical prediction: A theoretical analysis and a review of the evidence.

\bibitem[Meehl, 1986]{Meehl1986}
Meehl, P.~E. (1986).
\newblock Causes and effects of my disturbing little book.
\newblock {\em Journal of Personality Assessment}, 50(3):370--375.

\bibitem[Mendler-D{\"u}nner et~al., 2022]{MendlerDnner2022AnticipatingPB}
Mendler-D{\"u}nner, C., Ding, F., and Wang, Y. (2022).
\newblock Anticipating performativity by predicting from predictions.
\newblock In {\em Neural Information Processing Systems}.

\bibitem[Mendler-D{\"u}nner et~al., 2020]{MendlerDnner2020StochasticOF}
Mendler-D{\"u}nner, C., Perdomo, J.~C., Zrnic, T., and Hardt, M. (2020).
\newblock Stochastic optimization for performative prediction.
\newblock {\em ArXiv}, abs/2006.06887.

\bibitem[Mozannar and Sontag, 2020]{consistent-deferral-2020}
Mozannar, H. and Sontag, D.~A. (2020).
\newblock Consistent estimators for learning to defer to an expert.
\newblock In {\em International Conference on Machine Learning}.

\bibitem[Mullainathan and Obermeyer, 2019]{diagnosing-physician-error-2019}
Mullainathan, S. and Obermeyer, Z. (2019).
\newblock Diagnosing physician error: A machine learning approach to low-value health care.
\newblock Technical report, National Bureau of Economic Research, Cambridge, MA.

\bibitem[NEJM, 2018]{doi:10.1056/CAT.18.0194}
NEJM (2018).
\newblock Hospital readmissions reduction program (hrrp).
\newblock {\em Catalyst Carryover}, 4(2).

\bibitem[Oakland et~al., 2017]{OAKLAND2017635}
Oakland, K., Jairath, V., Uberoi, R., Guy, R., Ayaru, L., Mortensen, N., Murphy, M.~F., and Collins, G.~S. (2017).
\newblock Derivation and validation of a novel risk score for safe discharge after acute lower gastrointestinal bleeding: a modelling study.
\newblock {\em The Lancet Gastroenterology \& Hepatology}, 2(9):635--643.

\bibitem[Okati et~al., 2021]{differentiable-triage-2021}
Okati, N., De, A., and Gomez-Rodriguez, M. (2021).
\newblock Differentiable learning under triage.

\bibitem[Peery et~al., 2022]{PEERY2022621}
Peery, A.~F., Crockett, S.~D., Murphy, C.~C., Jensen, E.~T., Kim, H.~P., Egberg, M.~D., Lund, J.~L., Moon, A.~M., Pate, V., Barnes, E.~L., Schlusser, C.~L., Baron, T.~H., Shaheen, N.~J., and Sandler, R.~S. (2022).
\newblock Burden and cost of gastrointestinal, liver, and pancreatic diseases in the united states: Update 2021.
\newblock {\em Gastroenterology}, 162(2):621--644.

\bibitem[Perdomo et~al., 2020]{performative-prediction-2020}
Perdomo, J.~C., Zrnic, T., Mendler-D{\"u}nner, C., and Hardt, M. (2020).
\newblock Performative prediction.
\newblock In {\em Proceedings of the 37th International Conference on Machine Learning (ICML)}, volume abs/2002.06673.

\bibitem[Raghu et~al., 2019]{algo-triage-2019}
Raghu, M., Blumer, K., Corrado, G., Kleinberg, J., Obermeyer, Z., and Mullainathan, S. (2019).
\newblock The algorithmic automation problem: Prediction, triage, and human effort.

\bibitem[Rambachan, 2022]{prediction-mistakes-2022}
Rambachan, A. (2022).
\newblock Identifying prediction mistakes in observational data.

\bibitem[Rastogi et~al., 2022]{taxonomy-complementarity-2022}
Rastogi, C., Leqi, L., Holstein, K., and Heidari, H. (2022).
\newblock A taxonomy of human and ml strengths in decision-making to investigate human-ml complementarity.

\bibitem[Simon, 1957]{models-of-man-1957}
Simon, H.~A. (1957).
\newblock {\em Models of Man: Social and Rational}.
\newblock Wiley.

\bibitem[Stanley et~al., 2009]{Stanley2009-gq}
Stanley, A.~J., Ashley, D., Dalton, H.~R., Mowat, C., Gaya, D.~R., Thompson, E., Warshow, U., Groome, M., Cahill, A., Benson, G., Blatchford, O., and Murray, W. (2009).
\newblock Outpatient management of patients with low-risk upper-gastrointestinal haemorrhage: multicentre validation and prospective evaluation.
\newblock {\em Lancet}, 373(9657):42--47.

\bibitem[Stanley et~al., 2017]{Stanleyi6432}
Stanley, A.~J., Laine, L., Dalton, H.~R., Ngu, J.~H., Schultz, M., Abazi, R., Zakko, L., Thornton, S., Wilkinson, K., Khor, C. J.~L., Murray, I.~A., and Laursen, S.~B. (2017).
\newblock Comparison of risk scoring systems for patients presenting with upper gastrointestinal bleeding: international multicentre prospective study.
\newblock {\em BMJ}, 356.

\bibitem[Steyvers et~al., 2022]{bayesian-complementarity-2022}
Steyvers, M., Tejeda, H., Kerrigan, G., and Smyth, P. (2022).
\newblock Bayesian modeling of {human-AI} complementarity.
\newblock {\em Proc. Natl. Acad. Sci. U. S. A.}, 119(11):e2111547119.

\bibitem[Stolper et~al., 2011]{Stolper2011}
Stolper, E., Van~de Wiel, M., Van~Royen, P., Van~Bokhoven, M., Van~der Weijden, T., and Dinant, G.-J. (2011).
\newblock Gut feelings as a third track in general practitioners' diagnostic reasoning.
\newblock {\em Journal of General Internal Medicine}, 26(2):197--203.
\newblock Epub 2010 Oct 22.

\bibitem[Tversky and Kahneman, 1974]{heuristics-biases-1974}
Tversky, A. and Kahneman, D. (1974).
\newblock Judgment under uncertainty: Heuristics and biases.
\newblock {\em Science}, 185(4157):1124--1131.

\bibitem[Ur-Rahman et~al., 2018]{URRAHMAN2018}
Ur-Rahman, A., Guan, J., Khalid, S., Munaf, A., Sharbatji, M., Idrisov, E., He, X., Machavarapu, A., and Abusaada, K. (2018).
\newblock Both full glasgow-blatchford score and modified glasgow-blatchford score predict the need for intervention and mortality in patients with acute lower gastrointestinal bleeding.
\newblock {\em Digestive Diseases and Sciences}, 63:3020--3025.

\bibitem[van Amsterdam et~al., 2023]{harmful-prophecies-2023}
van Amsterdam, W., van Geloven, N., Krijthe, J.~H., Ranganath, R., and Cin'a, G. (2023).
\newblock When accurate prediction models yield harmful self-fulfilling prophecies.
\newblock {\em ArXiv}, abs/2312.01210.

\bibitem[Wainwright, 2019]{Wainwright2019-ii}
Wainwright, M.~J. (2019).
\newblock {\em High-dimensional statistics}.
\newblock Cambridge Series in Statistical and Probabilistic Mathematics. Cambridge University Press, Cambridge, England.

\bibitem[Wilder et~al., 2020]{learning-to-complement-2020}
Wilder, B., Horvitz, E., and Kamar, E. (2020).
\newblock Learning to complement humans.

\bibitem[Zhao et~al., 2023]{Zhao2023PerformativeTF}
Zhao, Z., Rodr{\'i}guez, A., and Prakash, B.~A. (2023).
\newblock Performative time-series forecasting.
\newblock {\em ArXiv}, abs/2310.06077.

\end{thebibliography}

\newpage 

\appendix

\section{Proof of \Cref{thm: optimality of linear regression}}
\label{sec: proof_thm_1}

\begin{proof}
A well known fact about univariate linear regression is that the coefficient of determination (or $r^2$) is equal to the square of the Pearson correlation coefficient between the regressor and the outcome (or $r$). In our context, this means that within any indistinguishable subset $S_k$ we have:

\begin{align}
    1 - \frac{\E_k\left[ \left(Y - \gamma^*_{k} - \beta^*_{k} \hY\right)^2 \right]}{\E_k\left[ \left(Y - \E_k[Y]\right)^2 \right]} = \frac{\Cov_k(Y, \hY)^2}{\Var_k(Y)\Var_k(\hY)}.
    \end{align}

This implies that
    \begin{align}
    \E_k \left[ \left(Y - \E_k[Y]\right)^2 \right] - \E_k \left[ \left(Y - \gamma^*_j - \beta^*_j \hY\right)^2 \right] & = \frac{\Cov_k(Y, \hY)^2}{\Var(\hY)}. \end{align}

Which finally implies that
\begin{align}
    \E_k \left[ \left(Y - \gamma^*_j - \beta^*_j \hY\right)^2 \right] & = \E_k \left[ \left(Y - \E_k[Y]\right)^2 \right]  - \frac{\Cov_k(Y, \hY)^2}{\Var(\hY)} \\
    & \le \E_k \left[ \left(Y - \E_k[Y]\right)^2 \right]  - 4 \Cov_k(Y, \hY)^2 \label{step: popoviciu, optimality of linear regression},
\end{align}

where \eqref{step: popoviciu, optimality of linear regression} is an application of Popoviciu's inequality for variances, and makes use of the fact that $\hY \in [0, 1]$ almost surely. We can then obtain the final result by applying the following lemma, which extends the main result in \citet{omnipredictors}. We provide a proof in \Cref{sec: aux lemmas}, but for now simply state the result as \Cref{lemma: omnipredictors extension} below.

\begin{lemma}
\label{lemma: omnipredictors extension}

Let $\{S_k\}_{k \in [K]}$ be an $\alpha$-multicalibrated partition with respect to a real-valued function class $\cF = \{ f: \cX \rightarrow [0, 1] \}$ and target outcome $Y \in [0, 1]$. For all $f \in \cF$ and $k \in [K]$, it follows that:
\begin{align}
\E_k\left[ \left(Y - \E_k[Y]\right)^2 \right] \le \E_k\left[ \left(Y - f(X)\right)^2 \right] + 2 \alpha \label{step: apply omnipredictors extension}
\end{align}
\end{lemma}

We provide further discussion of the relationship between \Cref{lemma: omnipredictors extension} and the main result of \citet{omnipredictors} in \Cref{sec: omnipredictors comparison} below.

Chaining inequalities \eqref{step: apply omnipredictors extension} and \eqref{step: popoviciu, optimality of linear regression} yields the final result:
\begin{align}
  \E_k \left[ \left(Y - \gamma^*_j - \beta^*_j \hY\right)^2 \right] \le \E_k \left[ \left(Y - f(X)\right)^2 \right] + 2 \alpha - 4\Cov_k(Y, \hY)^2 \hs{4pt} \forall \hs{4pt} f \in \cF\\
  \Rightarrow \E_k \left[ \left(Y - \gamma^*_j - \beta^*_j \hY\right)^2 \right] + 4\Cov_k(Y, \hY)^2 \le \E_k \left[ \left(Y - f(X)\right)^2 \right] + 2 \alpha \hs{4pt} \forall \hs{4pt} f \in \cF
\end{align}
 
\end{proof}

\section{Proof of \Cref{cor: high dimensional feedback}}
\label{sec: proof_cor_1}

\begin{proof}
    The proof is almost immediate. Let $\gamma^*$, $\beta^* \in \reals$ be the population regression coefficients obtained by regressing $Y$ on $g(H)$ within $S$ (as in \Cref{thm: optimality of linear regression}; the only difference is that we consider a single indistinguishable subset rather than a multicalibrated partition). This further implies, by the approximate calibration condition \eqref{eq: approx calibration}:

    \begin{align}
        \E_S \left[ (Y - g(H))^2 \right] \le \E_S \left[ (Y - \gamma_k^* - \beta_k^* g(H))^2\right] + \eta 
    \end{align}

    The proof then follows from that of \Cref{thm: optimality of linear regression}, replacing $\hY$ with $g(H)$. 
    
\end{proof}

\section{Proof of \Cref{thm: sufficiency of predictor implies bounded covariance}}
\label{sec: proof_thm_2}

\begin{proof} Fix any $k \in [K]$.

\begin{align}
     \left|\Cov_k(Y, \hY)\right| \\
     & = \left| \E_k[\Cov_k(Y, \hY \mid \tilde{f}_k(X)] + \Cov_k(\E_k[Y \mid \tilde{f}_k(X)], \E_k[\hY \mid \tilde{f}_k(X)]) \right| \label{step: law of total covariance, sufficiency lemma} \\
     & = \left| \Cov_k(\E[Y \mid \tilde{f}_k(X)], \E_k[\hY \mid \tilde{f}_k(X)]) \right| \label{step: Y,hY cond. independent, sufficiency lemma} \\
     & \le \sqrt{\Var(\E_k[Y \mid \tilde{f}_k(X)])\Var_k(\E[\hY \mid \tilde{f}_k(X)])} \label{step: cauchy-schwarz, sufficiency lemma} \\
     & \le \frac{1}{2}\sqrt{\Var_k(\E_k[Y \mid \tilde{f}_k(X)])} \label{step: popoviciu, sufficiency lemma}
\end{align}

Where \eqref{step: law of total covariance, sufficiency lemma} is the law of total covariance, \eqref{step: Y,hY cond. independent, sufficiency lemma} follows from the assumption that $Y \indep \hY \mid \tilde{f}_k(X), X \in S_k$, \eqref{step: cauchy-schwarz, sufficiency lemma} is the Cauchy-Schwarz inequality and \eqref{step: popoviciu, sufficiency lemma} applies Popoviciu's inequality to bound the variance of $\E[\hY \mid \tilde{f}_k(X)]$ (which is assumed to lie in $[0, 1]$ almost surely).

We now focus on bounding $\Var_k(\E_k[Y \mid \tilde{f}_k(X)])$. Recall that by assumption, $|\Cov_k(Y, \tilde{f}_k(X))| \le \alpha$, so we should expect that conditioning on $\tilde{f}_k(X)$ does not change the expectation of $Y$ by too much.

\begin{align}
    \Var_k & (\E_k[Y \mid \tilde{f}_k(X)]) \\
    & = \E_k[(\E_k[Y \mid \tilde{f}_k(X)] - \E_k[\E_k[Y \mid \tilde{f}_k(X)]])^2] \\
    & = \E_k[(\E_k[Y \mid \tilde{f}_k(X)] - \E_k[Y])^2]\\
    & = \p_k(\tilde{f}_k(X) = 1)(\E_k[Y \mid \tilde{f}_k(X) = 1] - \E_k[Y])^2 \nonumber \\
    & \qquad \qquad + \p_k(\tilde{f}_k(X) = 0)(\E_k[Y \mid \tilde{f}_k(X) = 0] - \E_k[Y])^2 \\
    & \le \p_k(\tilde{f}_k(X) = 1)\left|\E_k[Y \mid \tilde{f}_k(X) = 1] - \E_k[Y]\right| \nonumber \\
    & \qquad \qquad + \p_k(\tilde{f}_k(X) = 0)\left|\E_k[Y \mid \tilde{f}_k(X) = 0] - \E_k[Y]\right| \label{step: bounding squared difference, sufficiency lemma}
\end{align}

Where the last step follows because $Y$ is assumed to be bounded in $[0, 1]$ almost surely. We now make use of the following lemma:

\begin{lemma}\label{lemma: cov identity}
The following simple lemma will be useful in our subsequent proofs. Let $X \in \{0, 1\}$ be a binary random variable. Then for any other random variable $Y$, the following pair of equalities hold:

\begin{align}
    \Cov(X, Y) & = \p(X = 1)\left(\E[Y \mid X = 1] - \E[Y] \right) \label{eq: cov identity, X=1},
    \end{align}
    \begin{align}
    \Cov(X, Y) & = \p(X = 0)\left(\E[Y] - \E[Y \mid X = 0] \right) \label{eq: cov identity, X=0}.
\end{align}

This is exactly corollary 5.1 in \citet{omnipredictors}. We provide the proof in \Cref{sec: aux lemmas}.
    
\end{lemma}

Applying \Cref{lemma: cov identity} to \eqref{step: bounding squared difference, sufficiency lemma} yields

\begin{align}
    \Var_k(\E_k[Y \mid \tilde{f}_k(X)]) \le \left|2 \Cov_k(Y, \tilde{f}_k(X)) \right| \le 2\alpha \label{step: bounding variance of Y | f(X), sufficiency lemma},
\end{align}

where the second inequality follows because our analysis is conditional on $X \in S_k$ for some $\alpha$-indistinguishable subset $S_k$. Plugging \eqref{step: bounding variance of Y | f(X), sufficiency lemma} into \eqref{step: popoviciu, sufficiency lemma} completes the proof.
    
\end{proof}

\section{Additional technical results}
\label{sec:additional_tech}
In this section we present additional technical results which complement those in the main text. All proofs are deferred to \Cref{sec: additional_proofs}.

\subsection{A nonlinear analog of \Cref{thm: optimality of linear regression}}

Below we provide a simple extension of \Cref{thm: optimality of linear regression} from univariate linear regression to arbitrary univariate predictors of $Y$ given $\hY$.

\begin{corollary}\label{cor: nonlinear predictors}

Let $S$ be an $\alpha$-indistinguishable subset with respect to a model class $\cF$ and target $Y$. Let $g: [0, 1] \rightarrow [0, 1]$ be a function which satisfies the following approximate Bayes-optimality condition for $\eta \ge 0$:

\begin{align}
    \E_S[(Y - g(\hY))^2] \le \E_S[(Y - \E_S[Y \mid \hY])^2] + \eta. \label{eq: approx optimality}
\end{align}

Then, for any $f \in \cF$, 

\begin{align}
     \E_S \left[ (Y - g(\hY))^2  \right] + 4 \Cov_S(Y, \hY)^2 \le
\E_S\left[ \left(Y - f(X) \right)^2 \right] + 2 \alpha + \eta.
\end{align}

\end{corollary}

That is, any function $g$ which is nearly as accurate (in terms of squared error) as the univariate conditional expectation function $\E_S[Y \mid \hY]$ provides the same guarantee as in \Cref{thm: optimality of linear regression}. This conditional expectation function is exactly what e.g., a logistic regression of $Y$ on $\hY$ seeks to model. We provide a proof in \Cref{sec: additional_proofs}.

\subsection{A finite sample analog of \Cref{cor: high dimensional feedback}}
\label{sec: generalization argument}

For simplicity, the technical results in \Cref{sec: results} are presented in terms of population quantities. In this section we consider the empirical analogue of \Cref{cor: high dimensional feedback}, and provide a generalization argument which relates these empirical quantities to the corresponding population results in \Cref{sec: results}. We focus our attention on \Cref{cor: high dimensional feedback}, as the proof follows similarly for \Cref{thm: optimality of linear regression} and \Cref{cor: nonlinear predictors}.

Let ${\cal G}$ be some class of predictors mapping ${\cal H}$ to $[0, 1]$. We'll begin with the setup of \Cref{cor: high dimensional feedback}, with $S \subseteq \cX$ denoting a fixed, measurable subset of the input space (we'll generalize to a full partition $S_1 \dots S_K$ below). Further let $n_S \equiv \sum_{i=1}^n \mathbbm{1}(x_i \in S)$ denote the number of training examples which lie in the subset $S \subseteq \cX$, and $\{y_i, h_i\}_{i=1}^{n_S}$ denote i.i.d. samples from the unknown joint distribution over the random variables $(Y, H)$ conditional on the event that $X \in S$. Let $\hat{g}_S \equiv \argmin_{g \in {\cal G}} \frac{1}{n_S} \sum_{i=1}^{n_S} (y_i - g(h_i))^2$ denote the empirical risk minimizer within $S$. Our goal will be to show that if there exists some $g^*_S \in {\cal G}$ satisfying \eqref{eq: approx calibration}, then the empirical risk minimizer $\hat{g}_S$ also approximately satisfies \eqref{eq: approx calibration} with high probability. 

\begin{lemma}
\label{lemma: finite sample single subset}
    Let ${\cal L} = \{\ell_g : g \in {\cal G}\}$, where $\ell_g(x, h) \equiv (y - g(h))^2$, denote the class of squared loss functions indexed by $g \in {\cal G}$, and let $R_{n_S}({\cal L})$ denote the Rademacher complexity of ${\cal L}$. Let $P(S) \equiv \p(X \in S)$ denote the measure of $S$.  Then, for any $\delta \ge 0$, with probability 
    at least $(1 - e^{-\frac{n P(S) \delta^2}{4}})(1 - 2e^{-P(S)n})$, we have
    
    \begin{align}
        \E_S[\ell_{\hat{g}}] \le \E_S[\ell_{g^*}] + 4 R_{n_S}({\cal L}) + 2\delta, \quad \mbox{and}\quad n_S \geq n P(S)/2. 
    \end{align}

\end{lemma}

That is, if there exists some $g^* \in {\cal G}$ satisfying \eqref{eq: approx calibration}, then the empirical risk minimizer $\hat{g} \in {\cal G}$ within the subset $S$ also satisfies \eqref{eq: approx calibration} with high probability, up to an approximation error that depends on the Rademacher complexity of ${\cal L}$. For many natural classes of functions, including linear functions, the Rademacher complexity (1) tends to $0$ as $n \rightarrow \infty$ and (2) can be sharply bounded in finite samples (see e.g., Chapter 4 in \citet{Wainwright2019-ii}). We provide a proof of \Cref{lemma: finite sample single subset} in \Cref{sec: additional_proofs}.

\medskip
\noindent{\bf Extending \Cref{lemma: finite sample single subset} to a partition of $\cX$.} 
The result above is stated for a single subset $S$, and depends critically on the \emph{measure} of that subset $P(S) \equiv \p(X \in S)$. We now show this generalization argument can be extended to a partition of the input space $S_1 \dots S_K \subseteq \cX$ by arguing that \Cref{lemma: finite sample single subset} applies to ``most'' instances sampled from the marginal distribution over $\cX$. Specifically, we show that the probability that $X$ lies in \emph{any} subset $S$ with measure $P(S)$ approaching $0$ is a low probability event; for the remaining inputs, \Cref{lemma: finite sample single subset} applies.

\begin{corollary}
\label{cor: generalization for partition}
    Let $\{S_k\}_{k \in [K]}$ be a (not necessarily multicalibrated) partition of the input space, chosen independently of the training data, and let $n_k$ denote the number of training observations which lie in a given subset $S_k \subseteq \cX$. Then, for any $\epsilon > 0$ and $\delta \ge 0$, $X$ lies in a subset $S_k \subseteq \cX$ such that,

     \begin{align}
        \E_k[\ell_{\hat{g}}] \le \E_k[\ell_{g^*}] + 4 R_{n_k}({\cal L}) + 2\delta
    \end{align}

    with probability at least $(1-\epsilon)(1-e^{\frac{-n_k \epsilon \delta^2}{4K}})(1 - 2e^{-\frac{n_k \epsilon}{K}})$ over the distribution of the training data $\{x_i, y_i, \hat{y}_i\}_{i=1}^n$ and a test observation $(x_{n+1}, y_{n+1}, \hat{y}_{n+1})$.
\end{corollary}

The preceding corollary indicates that \Cref{lemma: finite sample single subset} also holds for a ``typical'' subset of the input space, replacing the dependence on the measure of any given subset $P(S)$ with a lower bound on the probability that a test observation $x_{n+1}$ lies in some subset whose measure is at least $\frac{\epsilon}{K}$. We provide a formal proof in \Cref{sec: additional_proofs}.

\section{Learning multicalibrated partitions}
\label{sec:learning partitions}

In this section we discuss two sets of conditions on $\cF$ which enable the efficient computation of multicalibrated partitions. An immediate implication of our first result is that any class of Lipschitz predictors induce a multicalibrated partition. All proofs are deferred to \Cref{sec: mc_proofs}.

\noindent
\textbf{Level sets of $\cF$ are multicalibrated.} Observe that one way in which \Cref{def: indistinguishable subset} is trivially satisfied (with $\alpha = 0$) is whenever every $f \in \cF$ is \emph{constant} within a subset $S \subseteq \cX$. We relax this insight as follows: if the variance of every $f \in \cF$ is bounded within $S$, then $S$ is approximately indistinguishable with respect to $\cF$.

\begin{lemma}
\label{lemma: bounded variance induces indistinguishability}

     Let $\cF$ be a class of predictors and $S \subseteq \cX$ be a subset of the input space. If:

     \begin{align}
         \sup_{f \in \cF} \Var(f(X) \mid X \in S) \le 4\alpha^2, \label{eq: bounded variance}
     \end{align}
     
     then $S$ is $\alpha$-indistinguishable with respect to $\cF$ and $Y$.
\end{lemma}

This result yields a natural corollary: the approximate level sets of $\cF$ (i.e., sets in which the range of every $f \in \cF$ is bounded) are approximately indistinguishable. We state this result formally as \Cref{corollary: approximate level sets are indistinguishable} below. We use exactly this approach to finding multicalibrated partitions in our study of a chest X-ray classification task in \Cref{sec:experiments}.

\begin{corollary}
    \label{corollary: approximate level sets are indistinguishable}

    Let $\cF$ be a class of predictors whose range is bounded within some $S \subseteq \cX$. That is, for all $f \in \cF$:

    \begin{align}
        \sup_{x \in S} f(x) - \inf_{x' \in S} f(x') \le 4 \alpha
    \end{align}
    
    Then $S$ is $\alpha$-indistinguishable with respect to $\cF$.
\end{corollary}

\Cref{lemma: bounded variance induces indistinguishability} also implies a simple algorithm for finding multicalibrated partitions when $\cF$ is Lipschitz with respect to some distance metric $d: \cX \times \cX \rightarrow \reals$: observations which are close under $d(\cdot, \cdot)$ are guaranteed to be approximately indistinguishable with respect to $\cF$. We state this result formally as \Cref{corollary: lipschitz predictors induce multicalibrated partitions} below.

\begin{corollary}
    \label{corollary: lipschitz predictors induce multicalibrated partitions}

    Let $\cF^{\text{Lip}(L, d)}$ be the set of $L$-Lipschitz functions with respect to some distance metric $d(\cdot, \cdot)$ on $\cX$. That is:

    \begin{align}
        \left|f(x) - f(x')\right| \le L d(x, x') \hspace{4pt} \forall \hspace{4pt} f \in \cF^{\text{Lip}(L, d)}
    \end{align}
    
    Let $\{S_k\}_{k \in K}$ for $K \subseteq \mathbb{N}$ be some ($4\alpha / L$)-net on $\cX$ with respect to $d(\cdot, \cdot)$. Then $\{S_k\}_{k \in K}$ is $\alpha$-multicalibrated with respect to $\cF^{\text{Lip}(L, d)}$.     
\end{corollary}

Proofs of the results above are provided in \Cref{sec: mc_proofs}.

\noindent
\textbf{Multicalibration via boosting.} Recent work by \citet{multicalibration-boosting-regression} demonstrates that multicalibration is closely related to \emph{boosting} over a function class $\cF$. In this section we first provide conditions, adapted from \citet{multicalibration-boosting-regression}, which imply that the level sets of a certain predictor $h: \cX \rightarrow [0, 1]$ are multicalibrated with respect to $\cF$; that is, the set $\{x \mid h(x) = v \}$ for every $v$ in the range of $h$ is approximately indistinguishable. We then discuss how these conditions yield a natural boosting algorithm for \emph{learning} a predictor $h$ which induces a multicalibrated partition. In the lemma below, we use $\cR(f)$ to denote the range of a function $f$.

\begin{lemma}
\label{lemma: mc via boosting}
     Let $\cF$ be a function class which is closed under affine transformations; i.e., $f \in \cF \Rightarrow a + bf \in \cF$ for all $a, b \in \reals$, and let $\tilde{\cF} = \{f \in \cF \mid \cR(f) \subseteq [0, 1] \}$. Let $Y \in [0,1]$ be the target outcome, and $h: \cX \rightarrow [0, 1]$ be some predictor with countable range $\cR(h) \subseteq [0, 1]$. If, for all $f \in \cF, v \in \cR(h)$:
    \begin{align}
        \E\left[(h(X) - Y)^2 - (f(X) - Y)^2 \mid h(X) = v\right] < \alpha^2, \label{eq: improved squared error}
    \end{align}
    
    then the level sets of $h$ are $(2\alpha)$-multicalibrated with respect to $\tilde{\cF}$ and $Y$. 
\end{lemma}

To interpret this result, observe that \eqref{eq: improved squared error} is the difference between the mean squared error of $f$ and the mean squared error of $h$ within each level set $S_v = \{ x \in \cX \mid h(x) = v \}$. Thus, if the best $f \in \cF$ fails to significantly improve on the squared error of $h$ within a given level set $S_v$, then $S_v$ is indistinguishable with respect to $\tilde{\cF}$ (which is merely $\cF$ restricted to functions that lie in $[0, 1]$). \citet{multicalibration-boosting-regression} give a boosting algorithm which, given a squared error regression oracle\footnote{Informally, a squared error regression oracle for $\cF$ is an algorithm which can efficiently output $\argmin_{f \in \cF} \E[(Y - f(X)]^2]$ for any distribution over $X,Y$. When the distribution is over a finite set of training data, this is equivalent to empirical risk minimization. We refer to \citet{multicalibration-boosting-regression} for additional details, including generalization arguments.} for $\cF$, outputs a predictor $h$ which satisfies \eqref{eq: improved squared error}. We make use of this algorithm in \Cref{sec:experiments} to learn a partition of the input space in a visual prediction task. Although the class we consider there (the class of shallow regression trees $\cF^{\text{RT}5}$) is not closed under affine transformations, boosting over this class captures the spirit of our main result: while no $f \in  \cF^{\text{RT}5}$ can improve accuracy within the level sets of $h$, humans provide additional predictive signal within three of them.

Taken together, the results in this section demonstrate that multicalibrated partitions can be efficiently computed for many natural classes of functions, which in turn enables the application of results in \Cref{sec: results}. 

\section{Omitted proofs from \Cref{sec:additional_tech}}\label{sec: additional_proofs}
\subsection{Proof of \Cref{cor: nonlinear predictors}}

\begin{proof}
    Observe that, because the conditional expectation function $\E_k[Y \mid \hY]$ minimizes squared error with respect to all univariate functions of $\hY$, we must have:

    \begin{align}
        \E_S \left[ (Y - \E_S[Y \mid \hY])^2 \right] \le \E_S \left[ (Y - \gamma^* - \beta^* \hY)^2 \right]
    \end{align}

    Where $\gamma^* \in \reals$, $\beta^* \in \reals$ are the population regression coefficients obtained by regression $Y$ on $g(H)$ as in \Cref{thm: optimality of linear regression}. This further implies, by the approximate Bayes-optimality condition \eqref{eq: approx optimality}:

    \begin{align}
        \E_S \left[ (Y - g(\hY))^2 \right] \le \E_S \left[ (Y - \gamma_k^* - \beta_k^* \hY)^2\right] + \eta 
    \end{align}

    The proof then follows immediately from that of \Cref{thm: optimality of linear regression}. 
    
\end{proof}

\subsection{Proof of \Cref{lemma: finite sample single subset}}

\begin{proof}  We will adopt the notation from the setup of \Cref{cor: high dimensional feedback}. Further let ${\cal G}$ be some class of predictors mapping ${\cal H}$ to $[0, 1]$, over which we seek to learn the mean-squared-error minimizing $g^*_S \in {\cal G}$ within some subset $S \subseteq {\cal X}$.

Let $n_S \equiv \sum_{i=1}^n \mathbbm{1}(x_i \in S)$ denote the number of training examples which lie in the subset $S$,  and let $\widehat{\E}_{S}[\ell_g] \equiv \frac{1}{n_S} \sum_{i = 1}^{n_S} (Y_i - g(X_i))^2$ denote the empirical loss incurred by some $g \in {\cal G}$ within $S$. Finally, let $\E_S[\ell_g] \equiv E[(Y - g(X))^2 \mid X \in S]$ denote the population analogue of $\widehat{\E}_S[\ell_g]$.

By Hoeffding's inequality we have:

\begin{align}
    \p\left(n_S \ge \frac{n P(S)}{2}\right) \ge \left(1 - 2e^{-P(S) n}\right) \label{eq: lower bound nS}
\end{align}

that is, $n_S$ is at least half its expectation with high probability. Let ${\cal L} = \{ \ell_g : g \in {\cal G}\}$, where $\ell_g(x,y) = (y - g(x))^2$, be the class of squared loss functions indexed by $g \in {\cal G}$. Let $R_{n_S}({\cal L})$ denote the Rademacher complexity of ${\cal L}$, which is defined as follows:

\begin{definition}{Rademacher Complexity}

For a fixed $n \in \mathbb{N}$, let $\epsilon_1 \dots \epsilon_n$ denote $n$ i.i.d. Rademacher random variables (recall that Rademacher random variables take values $1$ and $-1$ with equal probability $\frac{1}{2}$). Let $Z_1 \dots Z_n$ denote i.i.d. random variables taking values in some abstract domain ${\cal Z}$, and let ${\cal T}$ be a class of real-valued functions over the domain ${\cal Z}$. The Rademacher complexity of ${\cal T}$ is denoted by $R_n({\cal T})$, and is defined as follows:
\begin{align}
    R_n({\cal T}) = \E \left[ \sup_{t \in {\cal T}} \left| \frac{1}{n} \sum_{i=1}^n \epsilon_i t(Z_i) \right| \right]
\end{align}

Where the expectation is taken over both $\epsilon_1 \dots \epsilon_n$ and $Z_1 \dots Z_n$. Intuitively, the Rademacher complexity is the expected maximum correlation between some $t \in {\cal T}$ and the noise vector $\epsilon_1 \dots \epsilon_n$.
\end{definition}

We now make use of a standard uniform convergence result, which is stated in terms of the Rademacher complexity of a function class. We reproduce this theorem (lightly adapted to our notation) from the textbook treatment provided in \citet{Wainwright2019-ii} below:

\begin{theorem}{(adapted from \citet{Wainwright2019-ii})}
\label{thm: uniform convergence}
For any $b$-uniformly-bounded class of functions ${\cal T}$, any positive integer $n \geq 1$ and any scalar $\delta \geq 0$, we have:

\begin{align}
    \sup_{t \in {\cal T}} \left| \frac{1}{n} \sum_{i = 1}^n t(Z_i) - \E[t(Z_1)] \right| \leq 2 \mathcal{R}_n (\mathcal{T}) + \delta
\end{align}

with probability at least $1 - \exp \left( - \frac{n \delta^2}{2b^2} \right)$. 
\end{theorem}

Applying \Cref{thm: uniform convergence} (noting that ${\cal L}$ is uniformly bounded in $[0, 1]$) implies:

\begin{align}
    \sup_{g \in {\cal G}} \left|\widehat{\E}_S[\ell_g] - \E_S[\ell_g]\right| \le 2 R_{n_S}({\cal L}) + \delta \label{eq: uniform convergence}
\end{align}

with probability at least $1-e^{\frac{-n_S \delta^2}{2}}$. Finally, combining \eqref{eq: lower bound nS} with \eqref{eq: uniform convergence} further implies, for any $\delta \ge 0$,

\begin{align}
    \E_S[\ell_{\hat{g}}] \le \E_S[\ell_{g^*}] + 4 R_{n_S}({\cal L}) + 2\delta
\end{align}

with probability at least $(1 - e^{-\frac{n P(S) \delta^2}{4}})(1 - 2e^{-P(S)n})$, as desired.

\subsection{Proof of \Cref{cor: generalization for partition}}

\begin{proof}
    We'll adopt the notation from the setup of \Cref{lemma: finite sample single subset} and \Cref{cor: generalization for partition}. Observe that, if we were to take the union of every element of the partition $\{S_k\}_{k \in K}$ with measure $P(S) \equiv \p(X \in S_k) \le \frac{\epsilon}{K}$, the result would be a subset of the input space with measure at most $K \times \frac{\epsilon}{K} = \epsilon$. Note that this is merely an analytical device; we need not identify which subsets these are. 
    
    Thus, with probability at least $(1-\epsilon)$, a newly sampled test observation will lie in some element of the partition $\{S_k\}_{k \in [K]}$ with measure at least $\frac{\epsilon}{K}$. Conditional on this event, we can directly apply the result of \Cref{lemma: finite sample single subset}, plugging in a lower bound of $\frac{\epsilon}{K}$ for $P(S)$. This yields, for any $\epsilon > 0$ and $\delta \ge 0$, $X$ lies in a subset $S_k \subseteq \cX$ such that,

     \begin{align}
        \E_k[\ell_{\hat{g}}] \le \E_k[\ell_{g^*}] + 4 R_{n_k}({\cal L}) + 2\delta
    \end{align}

     with probability at least $(1-\epsilon)(1-e^{\frac{-n \epsilon \delta^2}{4K}})(1 - 2e^{-\frac{n \epsilon}{K}})$ over the distribution of the training data $\{x_i, y_i, \hat{y}_i\}_{i=1}^n$ and a test observation $(x_{n+1}, y_{n+1}, \hat{y}_{n+1})$, as desired.
\end{proof}

\end{proof}

\section{Omitted proofs from \Cref{sec:learning partitions}}\label{sec: mc_proofs}

\subsection{Proof of  \Cref{lemma: bounded variance induces indistinguishability}}

\begin{proof}
    We want to show $|\Cov(Y, f(X) \mid X \in S)| \le \alpha$ for all $f \in \cF$ and some $S$ such that $\sup_{f \in \cF} \Var(f(X) \mid X \in S) \le 4\alpha^2$.

    Fix any $f \in \cF$. We then have:
    
    \begin{align}
        & |\Cov(Y, f(X) \mid X \in S)| \\
        & \le \sqrt{\Var(Y \mid X \in S)\Var(f(X) \mid X \in S)} \label{step: cauchy-schwarz singleton} \\
        & \le \sqrt{\frac{1}{4} \times \Var(f(X) \mid X \in S)} \label{step: popoviciu's smoothness} \\
        & \le \sqrt{\frac{1}{4} \times 4\alpha^2} \label{step: smoothness assumption} \\
        & = \alpha
    \end{align}

Where \eqref{step: cauchy-schwarz singleton} is the Cauchy-Schwarz inequality, \eqref{step: popoviciu's smoothness} is Popoviciu's inequality and makes use of the fact that $Y$ is bounded in $[0, 1]$ by assumption, and \eqref{step: smoothness assumption} uses the assumption that $\sup_{f \in \cF} \Var(f(X) \mid X \in S) \le 4\alpha^2$.

\end{proof}

\subsection{Proof of \Cref{corollary: approximate level sets are indistinguishable}}
\begin{proof}
    We want to show that $\forall \hspace{4pt} f \in \cF$:

    \begin{align}
     |\Cov(Y, f(X) \mid X \in S)| \le \alpha
    \end{align}

    By assumption, $f(X)$ is bounded in a range of $4\alpha$ within $S$. From this it follows by Popoviciu's inequality for variances that $\forall \hspace{4pt} f \in \cF$:

        \begin{align}
        & \Var(f(X) \mid X \in S_j) \le \frac{(4\alpha)^2}{4} = 4\alpha^2
    \end{align}

    The proof then follows from \Cref{lemma: bounded variance induces indistinguishability}.
    
\end{proof}

\subsection{Proof of \Cref{corollary: lipschitz predictors induce multicalibrated partitions}}
\begin{proof}
    We want to show that $\forall \hspace{4pt} f \in \cF^{\text{Lip}(L, d)}, k \in K$:

    \begin{align}
     |\Cov_k(Y, f(X))| \le \alpha
    \end{align}
    
    Because $S_k$ is part of a $4\alpha/L$-net, there exists some $m \in [0, 1]$ such that $\p(f(X) \in [m, m+4\alpha] \mid X \in S_k) = 1$; that is, $f(X)$ is bounded almost surely in some interval of length $4\alpha$. From this it follows by Popoviciu's inequality for variances that $\forall \hspace{4pt} f \in \cF^{\text{Lip}(L, d)}, k \in K$:

        \begin{align}
        & \Var_k(f(X)) \le \frac{(4\alpha)^2}{4} = 4\alpha^2
    \end{align}

    The remainder of the proof follows from \Cref{lemma: bounded variance induces indistinguishability}.
    
\end{proof}

\subsection{Proof of \Cref{lemma: mc via boosting}}

\begin{proof}
    The result follows Lemma 3.3 and Lemma 6.8 in \citet{multicalibration-boosting-regression}. We provide a simplified proof below, adapted to our notation. We'll use $\E_v[\cdot]$ to denote the expectation conditional on the event that $\{ h(X) = v \}$ for each $v \in \cR(h)$. We use $\Cov_v(\cdot, \cdot)$ analogously.

Our proof will proceed in two steps. First we'll show that:
\begin{align}
    \forall v \in \cR(h), f \in \cF, \E_v[(h(X) - Y)^2 - (f(X) - Y)^2 ] < \alpha^2 \\
    \Rightarrow \E_v[f(X)(Y - v)] < \alpha \hspace{4pt} \forall \hspace{4pt} v \in \cR(h), f \in \tilde{\cF} \label{eq: no squared error improvement implies approx-mc}
\end{align}

This condition states that if there does not exist some $v$ in the range of $h$ where the best $f \in \cF$ improves on the squared error incurred by $h$ by more than $\alpha^2$, then the predictor $h(\cdot)$ is $\alpha$-multicalibrated in the sense of \citet{multicalibration-boosting-regression} with respect to the constrained class $\tilde{\cF}$. We then show that the level sets of a predictor $h(\cdot)$ which satisfies \eqref{eq: no squared error improvement implies approx-mc} form a multicalibrated partition (\Cref{def: multicalibrated partition}). That is:

\begin{align}
    \E_v[f(X)(Y - v)] \le \alpha \hspace{4pt} \forall v \in \cR(h), f \in \tilde{\cF} \hspace{4pt} \Rightarrow \Cov_v(f(X), Y) \le 2\alpha \hspace{4pt} \forall v \in \cR(h), f \in \tilde{\cF} \label{eq: approx mc implies cov mc}
\end{align}

That is, the level sets $S_v = \{ x \mid h(x) = v \}$ form a $(2\alpha)$-multicalibrated partition with respect to $\tilde{\cF}$.

First, we'll prove the contrapositive of \eqref{eq: no squared error improvement implies approx-mc}. This proof is adapted from that of Lemma 3.3 in \citet{multicalibration-boosting-regression}. Suppose there exists some $v \in \cR(h)$ and $f \in \tilde{\cF}$ such that 

\begin{align}
    \E_v[f(X)(Y - v)] \ge \alpha
\end{align}

Then there exists $f' \in \cF$ such that:

\begin{align}
    \E_v[(f'(X) - Y)^2 - (h(X) - Y)^2] \ge \alpha^2
\end{align}

Proof: let $\eta = \frac{\alpha}{\E_v[f(X)^2]}$ and $f' = v + \frac{\alpha}{\E_v[f(X)^2]} f(X) = v + \eta f(X)$. Then:

\begin{align}
    \E_v & \left[(h(X) - Y)^2 - (f'(X) - Y)^2\right] \\ 
    & = \E_v \left[ (v - Y)^2 - (v + \eta f(X) - Y)^2 \right] \\
    & = \E_v \left[ v^2 + Y^2 - 2Yv - v^2 - \eta^2 f(X)^2 - Y^2 - 2v \eta f(X) + 2vY + 2 \eta f(X) Y \right] \\
    & = \E_v \left[ 2 \eta f(X) \left(Y - v\right) - \eta^2 f(X)^2\right] \\
    & = \E_v \left[ 2 \eta f(X) \left(Y - v\right) \right] - \frac{\alpha^2}{\E_v[f(X)^2]} \\
    & \ge 2 \eta \alpha - \frac{\alpha^2}{\E_v[f(X)^2]} \\
    & = \frac{\alpha^2}{\E_v[f(X)^2]} \\
    & \ge \alpha^2
\end{align}

Where the last step follows because we took $f \in \tilde{\cF}$, the subset of the function class $\cF$ which only takes values in $[0, 1]$. This implies that if instead $\E_v[(f'(X) - Y)^2 - (h(X) - Y)^2] < \alpha^2$ for all $v \in \cR(h), f' \in \cF$, then $\E_v[f(X)(Y - v)] < \alpha$ for all $v \in \cR(h)$ and $f \in \tilde{\cF}$. Next we prove \eqref{eq: approx mc implies cov mc}; that is, $\E_v[f(X)(Y - v)] < \alpha$ for all $v \in \cR(h)$ and $f \in \tilde{\cF}$ implies $\left|\Cov_v(f(X), Y)\right| \le 2 \alpha$ for all $v \in \cR(h), f \in \tilde{\cF}$.

The proof is adapted from that of Lemma 6.8 in \citet{multicalibration-boosting-regression}; our proof differs beginning at \eqref{eq: penultimate step, boosting}. Fix some $f \in \tilde{\cF}$ and $v \in \cR(h)$. By assumption we have, for all $v \in \cR(h)$ and $f \in \tilde{\cF}$,
\begin{align}
    \E_v[f(X)(Y - v)] < \alpha  \label{eq: mc condition, boosting}
\end{align}

Then we can show:

\begin{align}
  \left| \Cov_v(f(X), Y) \right| & \\
  & = \left| \E_v[f(X)Y] - \E_v[f(X)]\E_v[Y] \right| \\
  & = \left| \E_v[f(X)Y] - \E_v[f(X)]\E_v[Y] + v \E_v[f(X)] - v\E_v[f(X)] \right| \\
  & = \left| \E_v[f(X)(Y - v)] + \E_v[f(X)](v - \E_v[Y]) \right| \\
  & \le \left| \E_v[f(X)(Y - v)]\right| + \left| \E_v[f(X)](v - \E_v[Y]) \right| \\
  & = \left| \E_v[f(X)(Y - v)]\right| + \left| \E_v[f(X)](\E_v[Y] - v) \right| \\
  & \le \alpha + \left| \E_v[f(X)](\E_v[Y] - v) \right| \label{eq: penultimate step, boosting}
\end{align}

Where the last step follows from the assumption \eqref{eq: mc condition, boosting}. Now, let $f'(X) \equiv \E_v[f(X)]$ be the constant function which takes the value $\E_v[f(X)]$. We can write \eqref{eq: penultimate step, boosting} as follows:

\begin{align}
    \alpha + \left| \E_v[f(X)](\E_v[Y] - v) \right| & = \alpha + \left| f'(X)(\E_v[Y] - v) \right| \\ 
    & = \alpha + \left| \E_v[f'(X)(Y - v)] \right| \label{eq: last step, boosting}
\end{align}

Because $\cF$ is closed under affine transformations, it contains all constant functions, and thus, $f'(X) \in \cF$. $\tilde{\cF}$, by definition, is the subset of $\cF$ whose range lies in $[0, 1]$. Because $f \in \tilde{\cF}$, it must be that $\E_v[f(X)] \in [0, 1]$ and thus, $f' \in \tilde{\cF}$. So, we can again invoke \eqref{eq: mc condition, boosting} to show:

\begin{align}
    \alpha + \left| \E_v[f'(X)(Y - v)] \right| \le 2 \alpha \label{eq: final bound, boosting}
\end{align}

Which completes the proof.

\end{proof}

\section{Relating \Cref{lemma: omnipredictors extension} to Omnipredictors \citep{omnipredictors}}
\label{sec: omnipredictors comparison}

In this section we compare \Cref{lemma: omnipredictors extension} to the main result of \citet{omnipredictors}. While the main result of \citet{omnipredictors} applies broadly to convex, Lipschitz loss functions, we focus on the special case of minimizing squared error. In this case, we show that \Cref{lemma: omnipredictors extension} extends the main result of \citet{omnipredictors} to cover real-valued outcomes under somewhat weaker and more natural conditions. We proceed in three steps: first, to provide a self-contained exposition, we state the result of \citet{omnipredictors} for real-valued outcomes in the special case of squared error (\Cref{lemma: omnipredictors main result} and \Cref{lemma: alternate extension to multiclass labels} below). Second, we derive a matching bound using \Cref{lemma: omnipredictors extension} (our result), which we do by demonstrating that the conditions of \Cref{lemma: alternate extension to multiclass labels} imply the conditions of \Cref{lemma: omnipredictors extension}. Finally, we show that \Cref{lemma: omnipredictors extension} applies in more generality than \Cref{lemma: alternate extension to multiclass labels}, under conditions which match those of \Cref{def: multicalibrated partition}.

We first state the main result of \citet{omnipredictors} (adapted to our notation) below, which holds for binary outcomes $Y \in \{0, 1\}$.\footnote{As discussed in \Cref{sec: introduction}, we also continue to elide the distinction between the `approximate' multicalibration of \citet{omnipredictors} and our focus on individual indistinguishable subsets. The results in this section can again be interpreted as holding for the `typical' element of an approximately multicalibrated partition.}

\begin{lemma}[Omnipredictors for binary outcomes, specialized to squared error (\citet{omnipredictors}, Theorem 6.3)]
\label{lemma: omnipredictors main result}

Let $S$ be a subset which is $\alpha$-indistinguishable with respect to a real-valued function class $\cF$ and a binary target outcome $Y \in \{0, 1\}$. Then, for all $f \in \cF$,

\begin{align}
    \E_S\left[ \left(Y - \E[Y]\right)^2 \right] \le \E_S\left[ \left(Y - f(X)\right)^2 \right] + 4 \alpha 
\end{align}
    
\end{lemma}

This result makes use of the fact that for any fixed $y \in [0, 1]$, the squared error function is $2$-Lipschitz with respect to $f(x)$ over the interval $[0, 1]$. This is similar to \Cref{lemma: omnipredictors extension}, but requires that $Y$ is binary-valued. In contrast, \Cref{lemma: omnipredictors extension} allows for real-valued $Y \in [0, 1]$, and gains a factor of $2$ on the RHS.\footnote{Note that \Cref{lemma: omnipredictors extension} also requires that each $f \in \cF$ takes values in $[0, 1]$, but this is without loss of generality when the outcome is bounded in $[0, 1]$; projecting each $f \in \cF$ onto $[0, 1]$ can only reduce squared error.} \citet{omnipredictors} provide an alternate extension of \Cref{lemma: omnipredictors main result} to bounded, real-valued $Y$, which we present below for comparison to \Cref{lemma: omnipredictors extension}.

\subsection{Extending \Cref{lemma: omnipredictors main result} to real-valued $Y$} Fix some $\epsilon > 0$, and let $B(\epsilon) = \{0, 1, 2 \dots \lfloor \frac{2}{\epsilon} \rfloor\}$. Let $\tilde{Y}$ be a random variable which represents a discretization of $Y$ into bins of size $\frac{\epsilon}{2}$. That is, $\tilde{Y} = \min_{b \in B(\epsilon)} \left| Y - \frac{b \epsilon}{2} \right|$. Let $\cR(\tilde{Y})$ denote the range of $\tilde{Y}$. Observe that the following holds for any function $g: \cX \rightarrow [0, 1]$:
\begin{align}
    \left| \E[(\tilde{Y} - g(X))^2] - \E[(Y - g(X))^2]\right| \le \epsilon  \label{eq: sq error is lipschitz}
\end{align}

Where \eqref{eq: sq error is lipschitz} follows because the function $(y - g(x))^2$ is $2$-Lipschitz with respect to $g(x)$ over $[0, 1]$ for all $y \in [0, 1]$. We now work with the discretization of $\tilde{Y}$, and provide an analogue to \Cref{lemma: omnipredictors main result} under a modified indistinguishability condition for discrete-valued $\tilde{Y}$, which we'll show is stronger than \Cref{def: indistinguishable subset}.

\begin{lemma}[Extending \Cref{lemma: omnipredictors main result} to real-valued $Y$ (\citet{omnipredictors}, adapted from Theorem 8.1)]
\label{lemma: alternate extension to multiclass labels}
Let $\cR(f)$ denote the range of a function $f$, and let $1(\cdot)$ denote the indicator function. Let $S$ be a subset of the input space $\cX$ which satisfies the following condition with respect to a function class $\cF$ and discretized target $\tilde{Y}$:

For all $f \in \cF$ and $\tilde{y} \in \cR(\tilde{Y})$, if:
\begin{align}
    \left| \Cov_S(1(\tilde{Y} = \tilde{y}), f(X)) \right| \le \alpha \label{eq: multiclass mc}
\end{align}

Then:

\begin{align}
    \E_S\left[ \left(\tilde{Y} - \E_S[\tilde{Y}]\right)^2 \right] \le \E_S\left[ \left(\tilde{Y} - f(X)\right)^2 \right] + 2 \left \lceil \frac{2}{\epsilon} \right\rceil \alpha \label{eq: mc bound}
\end{align}
    
\end{lemma}

To interpret this result, observe that \eqref{eq: mc bound} yields a bound which is similar to \Cref{lemma: omnipredictors main result} under a modified `pointwise' indistinguishability condition \eqref{eq: multiclass mc} for any discretization $\tilde{Y}$ of $Y$. Combining \eqref{eq: mc bound} with \eqref{eq: sq error is lipschitz} further implies:

\begin{align}
    \E_S\left[ \left(Y - \E_S[\tilde{Y}]\right)^2 \right] \le \E_S\left[ \left(Y - f(X)\right)^2 \right] + 2 \left \lceil \frac{2}{\epsilon} \right \rceil \alpha + 2\epsilon \label{eq: mc bound real-valued}
\end{align}

\subsection{Deriving \Cref{lemma: alternate extension to multiclass labels} using \Cref{lemma: omnipredictors extension}}

We show next that the `pointwise' condition \eqref{eq: multiclass mc} for $\alpha \ge 0$ implies our standard indistinguishability condition (\Cref{def: indistinguishable subset}) for $\alpha' = \left \lceil \frac{2}{\epsilon} \right \rceil \alpha$. This will allow us to apply \Cref{lemma: omnipredictors extension} to obtain a bound which is identical to \eqref{eq: mc bound real-valued}. Thus, we show that \Cref{lemma: omnipredictors extension} is at least as general as \Cref{lemma: alternate extension to multiclass labels}.

\begin{lemma}
\label{lemma: relating multiclass mc to standard}
Let $S$ be a subset satisfying \eqref{eq: multiclass mc}. Then, for all $f \in \cF$,

\begin{align}
    \left| \Cov_S(\tilde{Y}, f(X)) \right| \le \left \lceil \frac{2}{\epsilon} \right \rceil \alpha \label{eq: relating multiclass mc to standard}
\end{align}
    
\end{lemma}

We provide a proof in \Cref{sec: aux lemmas}. Thus, combining assumption \eqref{eq: multiclass mc} with \Cref{lemma: omnipredictors extension} and \eqref{eq: sq error is lipschitz} recovers a result which is identical to \Cref{lemma: alternate extension to multiclass labels}. That is, for all $f \in \cF$:

\begin{align}
    \left| \Cov_S(1(\tilde{Y} = \tilde{y}), f(X)) \right| & \le \alpha \\
    & \Rightarrow \left| \Cov_S(\tilde{Y}, f(X)) \right| \le \left \lceil \frac{2}{\epsilon} \right \rceil \alpha \label{eq: mc condition implication} \\
    & \Rightarrow \E_S\left[ \left(\tilde{Y} - \E_S[\tilde{Y}]\right)^2 \right] \le \E_S\left[ \left(\tilde{Y} - f(X)\right)^2 \right] + 2 \left \lceil \frac{2}{\epsilon} \right \rceil \alpha \label{eq: extension lemma implication} \\
    & \Rightarrow \E_S\left[ \left(Y - \E[\tilde{Y}]\right)^2 \right] \le \E_S\left[ \left(Y - f(X)\right)^2 \right] + 2 \left \lceil \frac{2}{\epsilon} \right \rceil \alpha + 2 \epsilon  \label{eq: sq error lipschitz implication}
\end{align}

Where \eqref{eq: mc condition implication} follows from \Cref{lemma: relating multiclass mc to standard}, \eqref{eq: extension lemma implication} follows from \Cref{lemma: omnipredictors extension} and \eqref{eq: sq error lipschitz implication} follows from \eqref{eq: sq error is lipschitz}.

\subsection{Extending \Cref{lemma: omnipredictors extension} beyond \Cref{lemma: alternate extension to multiclass labels}}

Finally, to show that \Cref{lemma: omnipredictors extension} extends \Cref{lemma: alternate extension to multiclass labels}, it suffices to provide a distribution over $f(X)$ for some $f \in \cF$ and a discrete-valued $\tilde{Y}$ taking $l \ge 1$ values such that \Cref{def: indistinguishable subset} is satisfied at level $\alpha \ge 0$, but \eqref{eq: multiclass mc} is not satisfied at $\alpha' = (\alpha / l)$ (though in fact that taking $\alpha' = \alpha$ also suffices for the following counterexample).

Consider the joint distribution in which the events $\{\tilde{Y} = 0, f(X) = 0\}$, $\{\tilde{Y} = \frac{1}{2}, f(X) = \frac{1}{2}\}$ and $\{\tilde{Y} = \frac{1}{2}, f(X) = 1\}$ occur with equal probability $\frac{1}{3}$ conditional on $\{ X \in S\}$ for some $S \subseteq \cX$. We suppress the conditioning event $\{ X \in S \}$ for clarity. Then:

\begin{align}
 \Cov(1(\tilde{Y} = 0), f(X)) = \p(\tilde{Y} = 1)\left(\E[f(X) \mid \tilde{Y} = 0] - \E[f(X)]\right) = -\frac{1}{6}
\end{align}

On the other hand we have:

\begin{align}
 \Cov(\tilde{Y}, f(X)) & = \E[\tilde{Y}f(X)] - \E[\tilde{Y}]\E[f(X)] \\
 & = \E[\tilde{Y} \E[f(X) \mid \tilde{Y}]] - \E[\tilde{Y}]\E[f(X)] \\
 & = \left(\frac{1}{3} \times 0 + \frac{2}{3} \times \frac{1}{2} \times \frac{3}{4} \right) - \frac{1}{3} \times \frac{1}{2} = \frac{1}{12}
\end{align}

That is, we have $\left| \Cov(\tilde{Y}, f(X)) \right| = \frac{1}{12} < 3\left| \Cov(1(\tilde{Y} = 0), f(X)) \right| = \frac{1}{2}$. Thus, \Cref{lemma: omnipredictors extension} establishes a result which is similar to \eqref{lemma: alternate extension to multiclass labels} for real-valued $Y$ under the weaker and more natural condition that $\left| \Cov(Y, f(X)) \right|$ is bounded, which remains well-defined for real-valued $Y$, rather than requiring the stronger pointwise bound \eqref{eq: multiclass mc} for some discretization $\tilde{Y}$.

Finally, we briefly compare \Cref{lemma: omnipredictors extension} to Theorem 8.3 in \citet{omnipredictors}, which generalizes \Cref{lemma: alternate extension to multiclass labels} to hold for linear combinations of the functions $f \in \cF$ and to further quantify the gap between the `canonical predictor' $\E_k[Y]$ and any $f \in \cF$ (or linear combinations thereof). These extensions are beyond the scope of our work, but we briefly remark that the apparently sharper bound of Theorem 8.3 is due to an incorrect assumption that the squared loss $(y - g(x))^2$ is $1$-Lipschitz with respect to $g(x)$ over the interval $[0, 1]$, for any $y \in [0, 1]$. Correcting this to a Lipschitz constant of $2$ recovers the same bound as \eqref{eq: sq error lipschitz implication}.

\section{Proofs of auxiliary lemmas}\label{sec: aux lemmas}

\subsection{Proof of \Cref{lemma: cov identity}}

\begin{proof}
We'll first prove \eqref{eq: cov identity, X=1}.
    \begin{align}
        \Cov(X, Y) & = \E[XY] - \E[X]\E[Y] \\ 
        & =  \E[\E[XY \mid X]] - \E[X]\E[Y]  \\
        & = \p(X = 1)\E[XY \mid X = 1] + \p(X = 0)\E[XY \mid X = 0] - \E[X]\E[Y] \\
        & =  \p(X = 1)\E[Y \mid X = 1] - \E[X]\E[Y] \\
        & = \p(X = 1)\E[Y \mid X = 1] - \p(X = 1)\E[Y] \\ 
        & = \p(X = 1) \left(\E[Y \mid X = 1] - \E[Y] \right)
    \end{align}

As desired. To prove \eqref{eq: cov identity, X=0}, let $X' = 1 - X$. Applying the prior result yields:

\begin{align}
    \Cov(X', Y) = \p(X' = 1) \left( \E[Y \mid X' = 1] - \E[Y] \right)
\end{align}
Because $X' = 1 \Leftrightarrow X = 0$, it follows that:

\begin{align}
    \Cov(X', Y) = \p(X = 0) \left( \E[Y \mid X = 0] - \E[Y] \right) \label{eq: cov identity}
\end{align}

Finally, because covariance is a bilinear function, $\Cov(X', Y) = \Cov(1-X,Y) = -\Cov(X,Y)$. Chaining this identity with \eqref{eq: cov identity} yields the result. 
\end{proof}

\subsection{Proof of \Cref{lemma: omnipredictors extension}}

The result we want to prove specializes Theorem 6.3 in \citet{omnipredictors} to the case of squared error, but our result allows $Y \in [0, 1]$ rather than $Y \in \{0, 1\}$. The first few steps of our proof thus follow that of Theorem 6.3 in \citet{omnipredictors}; our proof diverges starting at \eqref{step: omnipredictors, subtraction}. We provide a detailed comparison of these two results in \Cref{sec: omnipredictors comparison} above.

\begin{proof} Fix any $k \in [K]$. We want to prove the following bound:

\begin{align}
    \E_k[(Y - \E_k[Y])^2] \le \E_k[(Y - f(X))^2] + 4\alpha
\end{align}

It suffices to show instead that:

\begin{align}
    \E_k[(Y - \E_k[f(X)])^2] \le \E_k[(Y - f(X))^2] + 4\alpha
\end{align}

From this the result follows, as $\E_k[(Y - \E_k[Y])^2] \le \E_k[(Y - c)^2]$ for any constant $c$. To simplify notation, we drop the subscript $k$ and instead let the conditioning event $\{ X \in S_k \}$ be implicit throughout. We first show:

\begin{align}
    \E[(Y - f(X))^2] = \E \left[ \E \left[(Y - f(X))^2 \mid Y\right]\right] \ge \E \left[(Y - \E[f(X) \mid Y])^2\right]   \label{step: omnipredictors, jensen's bound}
\end{align}

Where the second inequality is an application of Jensen's inequality (the squared loss is convex in $f(X)$). From this it follows that:

\begin{align}
    \E\big[ & (Y - \E[f(X)])^2\big] - \E\left[(Y - f(X))^2\right] \\ 
    & \le \E \left[(Y - \E[f(X)])^2 - (Y - \E[f(X) \mid Y])^2\right] \label{step: omnipredictors, subtraction} \\
    & = \E \left[\E[f(X)]^2 -2Y\E[f(X)] - \E[f(X) \mid Y]^2 + 2Y\E[f(X) \mid Y] \right] \\
    & = 2\left(\E\left[Y\E[f(X) \mid Y] -Y\E[f(X)]  \right]\right) -\E\left[\E[f(X) \mid Y]^2 + \E[f(X)]^2\right] \\
    & = 2\left(\E\left[Yf(X)\right] -\E[Y]\E[f(X)]\right) -\E\left[\E[f(X) \mid Y]^2 + \E[f(X)]^2\right]  \\
    & = 2 \Cov(Y, f(X)) - \E\left[\E[f(X) \mid Y]^2\right] + \E[f(X)]^2 \\
    & = 2 \Cov(Y, f(X)) - \Var(\E[f(X) \mid Y]) \\
    & \le 2 \alpha \label{step: last step, omnipredictors ext}
\end{align}

Where each step until \eqref{step: last step, omnipredictors ext} follows by simply grouping terms and applying linearity of expectation. \eqref{step: last step, omnipredictors ext} follows by the multicalibration condition and the fact that the variance of any random variable is nonnegative.

\end{proof}

\subsection{Proof of \Cref{lemma: relating multiclass mc to standard}}
\begin{proof}
    Recall that $\tilde{Y}$ is a discrete random variable taking values $0, \frac{\epsilon}{2}, \epsilon, \frac{3\epsilon}{2} \dots \lfloor \frac{2}{\epsilon} \rfloor \frac{\epsilon}{2}$. We again use $\cR(\tilde{Y})$ to denote the range of $\tilde{Y}$. Our analysis below proceeds conditional on the event $\{ X \in S \}$, which we suppress for clarity. We can show

    \begin{align}
        \left| \Cov(\tilde{Y}, f(X)) \right| & = \left| \E[\tilde{Y}f(X)] - \E[\tilde{Y}]\E[f(X)] \right|\\
        & = \left| \E[\tilde{Y}f(X)] - \E[f(X)] \sum_{\tilde{y} \in \cR(\tilde{Y})} \tilde{y} \p(\tilde{Y} = \tilde{y}) \right| \\
        & = \left| \E[\tilde{Y} \E[f(X) \mid \tilde{Y}]] - \E[f(X)] \sum_{\tilde{y} \in \cR(\tilde{Y})}  \tilde{y} \p(\tilde{Y} = \tilde{y}) \right| \\
        & =  \left| \sum_{\tilde{y} \in \cR(\tilde{Y})}  \tilde{y} \p(\tilde{Y} = \tilde{y}) \E[f(X) \mid \tilde{Y} = \tilde{y}] - \E[f(X)] \sum_{\tilde{y} \in \cR(\tilde{Y})} \tilde{y} \p(\tilde{Y} = \tilde{y}) \right| \\
        & =  \left| \sum_{\tilde{y} \in \cR(\tilde{Y})}  \tilde{y} \p(\tilde{Y} = \tilde{y}) \left(\E[f(X) \mid \tilde{Y} = \tilde{y}] - \E[f(X)]\right) \right| \\
        & \le   \sum_{\tilde{y} \in \cR(\tilde{Y})}  \tilde{y} \p(\tilde{Y} = \tilde{y}) \left|  \left(\E[f(X) \mid \tilde{Y} = \tilde{y}] - \E[f(X)]\right) \right| \label{eq: y positive 1}\\ 
        & =   \sum_{\tilde{y} \in \cR(\tilde{Y})}  \tilde{y} \left|  \Cov(1(\tilde{Y} = \tilde{y}), f(X)) \right| \label{eq: mc cov identity} \\
        & \le   \sum_{\tilde{y} \in \cR(\tilde{Y})}  \tilde{y} \alpha \label{eq: applying mc cov bound}  \\ 
        & \le   \sum_{\tilde{y} \in \cR(\tilde{Y})} \alpha \label{eq: y bounded above} \\ 
        & \le  \left \lceil \frac{2}{\epsilon} \right \rceil \alpha 
    \end{align}

Where \eqref{eq: y positive 1} makes use of the fact that $\tilde{y} \ge 0$, \eqref{eq: mc cov identity} makes use of the identity $\left|\Cov(1(\tilde{Y} = \tilde{y}), f(X))\right| =  \p(\tilde{Y} = \tilde{y}) \left|  \left(\E[f(X) \mid \tilde{Y} = \tilde{y}] - \E[f(X)]\right) \right|$ (this is a straightforward analogue of \Cref{lemma: cov identity}), \eqref{eq: applying mc cov bound} applies assumption \eqref{eq: multiclass mc}, and \eqref{eq: y bounded above} makes use of the fact that $\tilde{y} \le 1$.

\end{proof}

\end{document}